\documentclass[10pt, a4paper]{article}

\usepackage[final]{lrec2026} % this is the new style
\usepackage{makecell}
\usepackage{booktabs}
\usepackage[most]{tcolorbox}
\usepackage{lipsum}
\usepackage{graphicx}
\usepackage{hyperref}
\usepackage{adjustbox}
\usepackage{rotating}
\usepackage{float} % in the preamble
\title{MTEB-NL and E5-NL: Embedding Benchmark and Models  for Dutch}

\name{\begin{tabular}{c}
    Nikolay Banar\textsuperscript{*}, Ehsan Lotfi\textsuperscript{*}, Jens Van Nooten \\
    Cristina Arhiliuc, Marija Kliocaite, Walter Daelemans
  \end{tabular}} 

\address{University of Antwerp \\
         Antwerp, Belgium \\
         \{nicolae.banari, ehsan.lotfi, walter.daelemans\}@uantwerpen.be\\
         }

\abstract{
Recently, embedding resources, including models, benchmarks, and datasets, have been widely released to support a variety of languages. However, the Dutch language remains underrepresented, typically comprising only a small fraction of the published multilingual resources. To address this gap and encourage the further development of Dutch embeddings, we introduce new resources for their evaluation and generation. First, we introduce the Massive Text Embedding Benchmark for Dutch (MTEB-NL), which includes both existing Dutch datasets and newly created ones, covering a wide range of tasks. Second, we provide a training dataset compiled from available Dutch retrieval datasets, complemented with synthetic data generated by large language models to expand task coverage beyond retrieval. Finally, we release a series of E5-NL models compact yet efficient embedding models that demonstrate strong performance across multiple tasks. We make our resources publicly available through the Hugging Face Hub and the MTEB package. 
 \\ \newline  }

\begin{document}

\maketitleabstract

\begingroup
\renewcommand\thefootnote{} % suppress number
\footnotetext{\textsuperscript{*} Equal contribution}
% \addtocounter{footnote}{-1} % keep counter correct
\endgroup

\section{Introduction}

Embedding models have recently achieved substantial progress, driven by advances in the development of large language models (LLMs) and the increasing availability of resources for training and evaluation \citep{zhao2024dense}. These models can either be initialized from existing LLMs or undergo full-scale pretraining, in both cases followed by a final fine-tuning stage on more task-specific datasets such as MS MARCO \citep{bajaj2016ms}, NQ \citep{kwiatkowski-etal-2019-natural}, HotpotQA \citep{yang2018hotpotqa} and others. As a result, generated vector representations encode rich semantic relationships and achieve strong performance in many tasks \citep{muennighoff2023mteb}, including retrieval, classification, and clustering. Despite these advances, Dutch embedding development still lags behind, as the language remains only moderately represented in multilingual embedding resources. Below, we highlight the underlying issues and present our contributions.

The first step in model development is the creation of reliable benchmarks and evaluation sets, enabling researchers to assess model performance, identify weaknesses, and guide future improvements. The Massive Text Embedding Benchmark (MTEB; \citealp{muennighoff2023mteb}) provides a convenient package to evaluate text embeddings in a zero-shot setting with minimal user effort. It covers a wide range of publicly available datasets and tasks, including retrieval, reranking, classification, and more.
Later on, this initiative was extended to multiple languages \citep{c-pack-2023-bge,ciancone2024mteb,poswiata2024pl,snegirev2024russian,enevoldsen2024scandinavian,wehrli2024germantextembeddingclustering,zinvandi2025famteb}. Building on this benchmarking suite, we compile and create Dutch resources, MTEB-NL, corresponding to the task categories from MTEB.

A limitation for building Dutch embedding models is the lack of fine-tuning datasets that can enrich the embedding space of these models. Until recently, the only large dataset available was the Dutch part of mMARCO \citep{bonifacio2021mmarco}, a multilingual translation of MS MARCO \citep{bajaj2016ms}. More recently, BEIR (Benchmarking IR; \citealp{thakur2beir}), a diverse and heterogeneous collection of retrieval datasets, was automatically translated to Dutch as BEIR-NL \citep{lotfi2025beir}. While BEIR was primarily used as a standard benchmark for zero-shot evaluation, its large training splits have made it a common resource for training embedding models. However, large retrieval datasets alone are not sufficient for building a strong general-purpose embedding model, as demonstrated by \citet{wang-etal-2024-improving-text}, \citet{choi2024linqembedmistraltechnicalreport}, and \citet{ DBLP:conf/iclr/Lee0XRSCP25}.
Inspired by their work, we construct a training mixture that combines available Dutch retrieval datasets with synthetic data generated by LLMs to expand task coverage.

Finally, using the developed resources for training and evaluation, we build our Dutch embedding models based on the E5 family \citep{wang2024multilingual}, compact yet robust models that achieve top performance on multilingual benchmarks \citep{enevoldsenmmteb}. Our models achieve state-of-the-art results among available Dutch embedding models and demonstrate strong performance in Dutch compared to multilingual ones.

Our contributions are as follows: (i) we release MTEB-NL, a massive text embedding benchmark for Dutch comprising 40 datasets, and evaluate existing models from MTEB on Dutch tasks; (ii) we curate a mixture of human-annotated and synthetic data for training general-purpose Dutch embedding models; (iii) we train a small suite of state-of-the-art Dutch embedding models.  We make all these resources available on the Hugging Face Hub\footnote{\url{https://huggingface.co/collections/clips/mteb-nl-6888d7136112c731605f93ed} and \url{https://huggingface.co/collections/clips/e5-nl-68be9d3760240ce5c7d9f831}} and  GitHub\footnote{\url{https://github.com/nikolay-banar/mteb-nl-dev} and \url{https://github.com/ELotfi/e5-nl}} to ensure easy access and reusability.

\section{Related Work}
Our related work is organized into three subsections reflecting our contributions. We first review existing embedding benchmarks, then discuss embedding models, and finally address synthetic embedding data.

\subsection{Embedding Benchmarks}

Since its creation, MTEB \citep{muennighoff2023mteb}, originally presented in English, has been extended to a multilingual context in MMTEB \citep{enevoldsenmmteb} as well as multiple individual languages. These initiatives generally fall into two categories: (i) compiling human-annotated datasets into benchmarks, and (ii) automatically translating existing benchmarks into new languages. Human-annotated data is typically of high quality but demands substantial time and financial resources, whereas machine-translated data is faster and more cost-effective in creation but offers lower quality for evaluations \cite{englander2024m2qa}.

\citet{c-pack-2023-bge} expanded MTEB to Chinese (C-MTEB) by gathering 35 publicly available datasets. MTEB-French \cite{ciancone2024mteb} contributed 18 datasets in French, drawing from both original resources and translations produced with DeepL. For German, \citet{wehrli2024germantextembeddingclustering} curated six datasets designed to benchmark clustering tasks. The Polish benchmark, MTEB-PL \cite{poswiata2024pl}, offers 28 datasets; its retrieval portion is based on BEIR-PL \cite{wojtasik2024beir}, a translation of a subset of BEIR into Polish via Google Translate. ruMTEB \cite{snegirev2024russian} comprises 23 tasks following the MTEB format, mainly original Russian datasets with one translated using DeepL. The Scandinavian Embedding Benchmark (SEB; \citealp{enevoldsen2024scandinavian}) provides 24 tasks across Scandinavian languages, combining native datasets with translated ones from MTEB. \citet{zinvandi2025famteb} present FaMTEB, a benchmark comprising 63 datasets for Persian, constructed from a mix of existing resources, translated corpora, and newly created data. Finally, MMTEB broadened the scope to more than 250 languages, covering  over  550 quality-controlled evaluation tasks. Compared to previous work, this benchmark suite emphasizes quality control by excluding machine-translated datasets, while also encouraging community contributions of smaller, high-quality resources.

% This collection was then reduced by filtering out machine-translated datasets, resources with unclear licenses, and domain-specific datasets such as code retrieval, resulting in 343 tasks across more than 250 languages.

The Dutch benchmarking landscape is scarce.  DUMB (A Benchmark for Smart Evaluation of Dutch Models; \citealp{de2023dumb}) consists of 9 diverse datasets. Although it provides open-access code, many of its datasets have restrictive licenses and cannot be publicly distributed. EuroEval \citep{smart2023scandeval, smart2024encoder} includes a subset of four datasets for Dutch embedding models, which mostly overlap with DUMB. Another Dutch benchmark, BEIR-NL \citep{lotfi2025beir} is focused on retrieval and comprised of many datasets extensively overused to fine-tune embedding models. 

Building on the previous efforts of the Dutch NLP community, we construct MTEB-NL following, as far as possible, the same principles used in the most recent update of MTEB, namely MMTEB \citep{enevoldsenmmteb}, with an aim to strike a balance between quality and coverage.

\subsection{Text Embedding Models}

The development of text embeddings (i.e. low-dimensional vector representations of text) has a rich history aimed at moving beyond sparse, high-dimensional representations like TF-IDF. Early works include methods like Latent Semantic Indexing (LSA) and Latent Dirichlet Allocation (LDA), with simpler approaches such as a weighted average of word vectors also proving to be strong baselines \cite{Wang2022TextEB}. 

The advent of pre-trained language models (PLMs) like BERT \cite{devlin-etal-2019-bert} and GPT \cite{Radford2018ImprovingLU} provided a straightforward way to produce dense text embeddings; however, it was soon realized that the raw output of these models were not optimal for similarity tasks, which led to the development of various fine-tuning techniques, initially focused on supervised fine-tuning using large-scale labeled datasets, such as SNLI, and leading to influential models like Sentence-BERT \cite{reimers-gurevych-2019-sentence}. SimCSE \cite{gao-etal-2021-simcse} and its variations showed that unsupervised contrastive learning can achieve competitive results without the need for labeled data. LaBSE \cite{feng-etal-2022-language} extended this idea to learn multilingual representations from parallel sentences.

For text retrieval tasks, which often involve an asymmetric relationship between a short query and a longer document, a line of research has focused on self-supervised pre-training to automatically generate vast numbers of training pairs. For example, Contriever \cite{izacard2022unsupervised} demonstrated that random cropping of passages can be used as queries to train effective retrieval models. While these self-supervised methods provide abundant training signals, the synthetically generated data can be of low quality, and models trained on them often require further fine-tuning on labeled data to outperform classic retrieval baselines like BM25. This multi-stage training paradigm (pre-training  on a massive corpus of weakly-supervised text pairs using a contrastive loss, followed by a fine-tuning stage on smaller, high-quality labeled datasets) has produced a new generation of powerful, general-purpose text embedding models such as GTE \cite{li2023generaltextembeddingsmultistage}, E5 \cite{Wang2022TextEB}, and BGE \cite{bge-m3}, followed by their multilingual versions \cite{zhang2024mgte, wang2024multilingual}. Although these multilingual models achieve high performance across a wide range of languages, the majority of their training data comes from English or Chinese sources, and multiple studies have tried to improve their results for less resourced languages, like Russian \cite{snegirev2024russian} and Arabic \cite{bhatia-etal-2025-swan}.      

In recent years the goal has increasingly shifted towards creating unified text representation models that perform well beyond retrieval and across a wider array of embedding tasks, including clustering, classification, and semantic text similarity. \citet{wang-etal-2024-improving-text} showed that using well-crafted synthetic data can effectively boost the performance of retrieval models on other tasks, thus establishing a mixed-data recipe that remains popular. 

\subsection{Synthetic Embedding Data} 
The use of synthetic data for training embedding models has been widely studied and practiced, either to augment real data or to replace it. Early examples were limited to retrieval, and used few-shot prompting with language models to generate questions from real documents \cite{inpars1-2022,jeronymo2023inparsv2largelanguagemodels}. \citet{lee2024geckoversatiletextembeddings} extended this to the more general problem of text embedding, and generated tasks as well as queries for a given real document, which was also adopted by \citet{merrick2024arcticembedscalableefficientaccurate} . Later \citet{wang-etal-2024-improving-text} leveraged LLMs in a two-step process to come up with various text embedding tasks, as well as full triplets (query, positive and negative document). Using these data in addition to existing QA datasets to train a decoder-only model, they achieved a performance comparable to previously used multi-stage training pipelines, and provided a successful recipe which was modified and improved to address similar situations \cite{bge-m3,Liu_Meng_2024,choi2024linqembedmistraltechnicalreport, DBLP:conf/iclr/Lee0XRSCP25}. More recently \citet{kim-baek-2025-syntriever} employed LLMs in a more extensive way to check and verify their generated triplets, in addition to generate document preference signals which they used for a second stage of alignment training on the retriever. Finally \citet{chen-etal-2025-little} finetuned and aligned small open-source LLMs to efficiently generate large-scale synthetic embedding data.

\section{MTEB-NL Benchmark}
\begin{figure*}[h]
    \centering
    \includegraphics[width=1\linewidth]{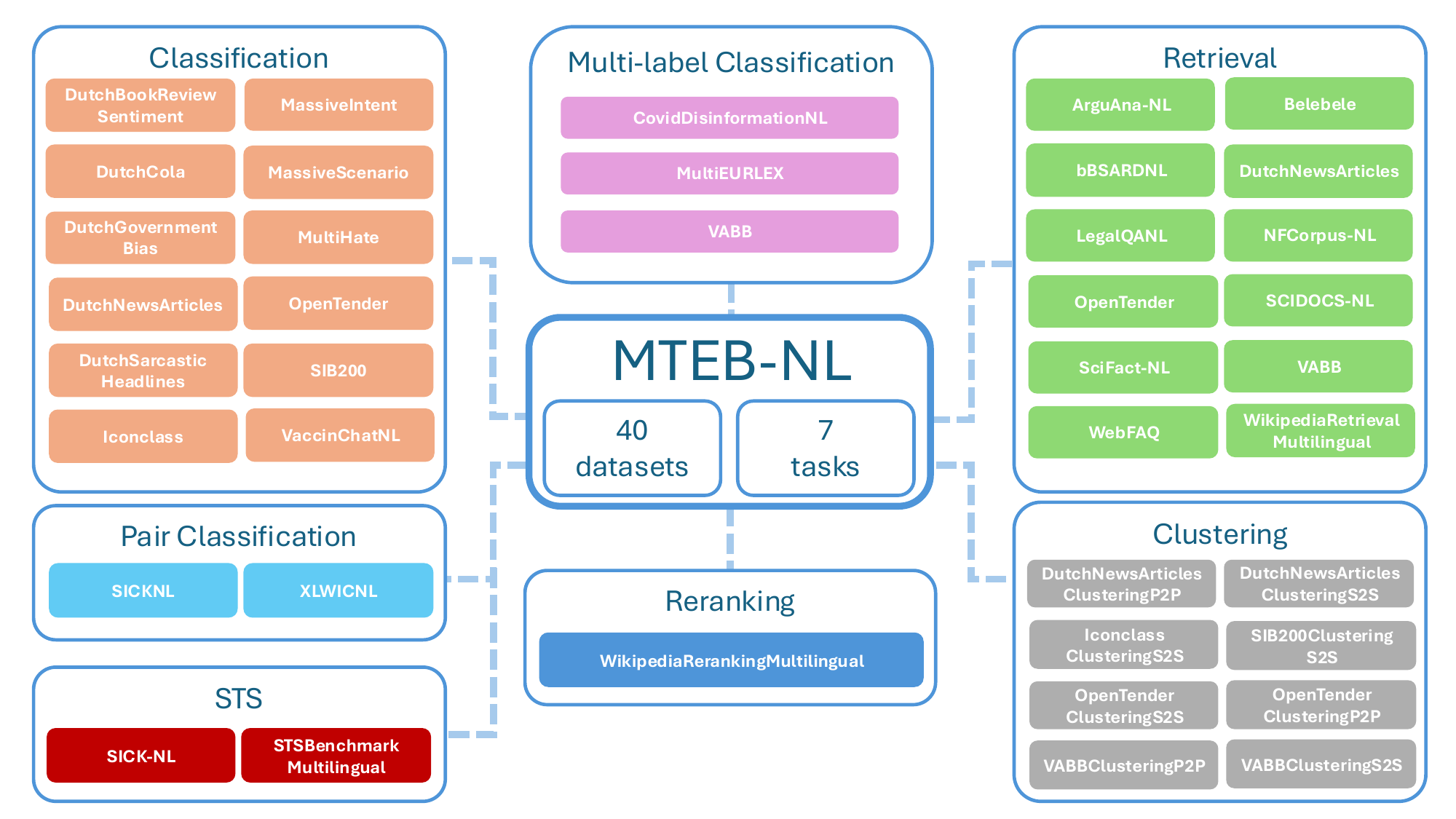} % or plot.png
    \caption{An overview of tasks and datasets in MTEB-NL.}
    \label{fig:mteb}
\end{figure*}

As mentioned before, we construct MTEB-NL following, as far as possible, the same principles as MTEB in its multilingual update. First, we select datasets that have rarely, if ever, been used during the fine-tuning of popular embedding models, in order to remain consistent with the zero-shot setting of MTEB. Second, we select or build compact datasets to avoid high computational costs. Third, we aim to reduce the proportion of machine-translated datasets in the benchmark. However, this latter goal is difficult to achieve given the scarcity of Dutch datasets. Finally, we select datasets with permissive licenses or those already publicly released by their authors. Figure \ref{fig:mteb} provides an overview of the datasets we collected and created. In total, we cover seven tasks from MTEB: classification, multi-label classification, pair classification, reranking, retrieval, clustering, and semantic textual similarity (STS).

We utilize 12 datasets already included in the MTEB package (excluding BEIR-NL): five for classification (DutchBookReviewSentimentClassification, MassiveIntentClassification, MassiveScenarioClassification, MultiHateClassification, SIB200Classification)
one for multi-label classification (MultiEURLEXMultilabelClassification), three for retrieval (BelebeleRetrieval, WebFAQRetrieval, WikipediaRetrievalMultilingual), one for reranking (WikipediaRerankingMultilingual), one for clustering (SIB200ClusteringS2S), and one for semantic textual similarity (STSBenchmarkMultilingualSTS). Although the latter is machine-translated, since STS resources are rarely available in Dutch, we include it.

Retrieval resources are also rarely available in Dutch. Although machine-translated, BEIR-NL could have served as a valuable candidate to cover the retrieval component of our benchmark. However, it includes large datasets, such as NQ \citep{kwiatkowski-etal-2019-natural}, FEVER \citep{thorne2018fever},  and HotpotQA \citep{yang2018hotpotqa}, that are extensively used for fine-tuning models. Many recent embedding models are exposed to different datasets from BEIR \citep{wang2024multilingual, wang-etal-2024-improving-text, DBLP:conf/iclr/Lee0XRSCP25,choi2024linqembedmistraltechnicalreport}. At the end, we select four small datasets (out of 14) from BEIR-NL \citep{lotfi2025beir} that are rarely used in fune-tuning and provide unique value for Dutch: ArguAna-NL, NFCorpus-NL, SCIDOCS-NL, SciFact-NL. The first two datasets originate from the medical domain, while the latter ones are sourced from the scientific domain.

Another benchmarking resource we investigated in greater detail is DUMB \citep{de2023dumb}. Some datasets do not fit the definition of MTEB tasks, such as Lassy (part-of-speech tagging; \citealp{van2012large}), SoNaR-1 (named entity recognition; \citealp{oostdijk2012construction}), and COPA-NL (causal reasoning; \citealp{de2023dumb}). Others have restrictive distribution terms, including Dutch Pronoun Resolution \citep{de2023dumb} and DALC v2.0 \citep{ruitenbeek-etal-2022-zo}. For certain datasets, suitable alternatives are already available, such as the Book Reviews Dataset (already included in MTEB; \citealp{van2019merits}), WiC-NL \citep{de2023dumb} (covered by XLWICNLPairClassification), and SQuAD-NL \citep{de2023dumb} (covered by WikipediaRetrievalMultilingual). We use SICK-NL \citep{wijnholds-moortgat-2021-sick} from this benchmark to enrich the STS and pair classification tasks.

Other datasets are drawn from existing resources or constructed by us. A detailed description of all datasets and preprocessing steps is provided in Appendix A. Table~\ref{tab:dutch_datasets} lists all datasets used, along with their citations and corresponding Hugging Face URLs. Table~\ref{tab:dutch_datasets_stat} summarizes dataset statistics, while Figure~\ref{fig:similarity} illustrates the similarity of documents in our benchmark.

\section{Dutch Embedding Models}
In this section, we describe the process of producing our suite of Dutch embedding models, based on the English and multilingual encoder-based E5 collections \cite{Wang2022TextEB,wang2024multilingual}. First, we go through the data curation step, and then we explain how these datasets are used to fine-tune the models.

\subsection{Training Data}
Following \citet{wang-etal-2024-improving-text}, \citet{choi2024linqembedmistraltechnicalreport}, and \citet{DBLP:conf/iclr/Lee0XRSCP25} we leverage a mixture of existing (human annotated) and synthetic datasets to train our embedding models.  Table \ref{tab:train_datasets_stat} shows an overview of the final training data. 

\subsubsection{Existing Datasets}
\label{existing-data}
The human annotated part includes the training set of the three commonly used retrieval datasets: mMARCO-NL \cite{bonifacio2021mmarco} (sampled by 75\%), FEVER-NL, and HotPotQA-NL \cite{lotfi2025beir}, which together amount to 620K samples. We mine hard negatives using multilingual-e5-large-instruct as the teacher, with a sampling method inspired by \citet{moreira2025nvretrieverimprovingtextembedding}, which we call TopK-STDMarginPos: for each sample, we rank the corpus using the teacher model, calculate the standard deviation of the top 1000 scores ($\sigma$), and use it as the ignore margin; i.e. when sampling, we skip the documents with scores between $S(d^{+})$ and $S(d^{+}) - \sigma$ where $S(d^{+})$ is the score of the positive document. This reduces the risk of including false negatives.

\begin{table}[ht]
\centering
\small
% \resizebox{1\textwidth}{!}{ 
\begin{tabular}{llcc}
% {@{}p{.25\textwidth} p{.20\textwidth} p{.40\textwidth} p{.11\textwidth} p{.11\textwidth}@{}}
\toprule
& \textbf{Dataset} & \textbf{Task} & \textbf{Size}  \\ 
\midrule
\textbf{Public} &&& \\
\midrule
& mMARCO-NL & Retrieval & 310K \\
& FEVER-NL & Retrieval & 140K \\
& HotPotQA-NL & Retrieval & 170K \\
\midrule
\textbf{Synthetic} &&& \\
\midrule
& short-long & Retrieval & 80K \\
& long-short & Classification & 140K \\
& short-short & Clustering & 15K \\
& long-long & Clustering & 15K \\
& STS & STS & 80K \\
\midrule
Total &&& 950K \\
\bottomrule 

\end{tabular}

\caption{Overview of the training data}
\label{tab:train_datasets_stat}
\end{table}

\subsubsection{Synthetic Data}
\label{synth-data}
As mentioned before, recently synthetic embedding data has been successfully used to enrich the task and domain coverage of human annotated datasets. The most common way to produce these data, is prompting an LLM with specific descriptions and instructions for each of the 5 main categories of embedding data, i.e. short-short, short-long, long-short, long-long and STS \cite{wang-etal-2024-improving-text}. However, the LLM-generated triplets (query, positive document, hard negative document) are prone to issues such as limited topic diversity and easy hard negatives \cite{choi2024linqembedmistraltechnicalreport, chen-etal-2025-little}. To address these issues, we employ 3 control methods: explicit topic sampling, instruction tuning, and data filtering, which we describe below.
\paragraph{Explicit Topic Sampling}
In the detailed study of the synthetic data strategy employed by \citet{wang-etal-2024-improving-text}, \citet{choi2024linqembedmistraltechnicalreport} identify content repetition and low diversity as the most common issue across all 5 task categories. They propose techniques such as few-shot instruction and prompt engineering (e.g. encouraging the model to use multiple entities or colloquial language) to mitigate the problem. \citet{chen-etal-2025-little} address this issue more directly by sampling topics from the Open Directory Project\footnote{\url{http://odp.org}}, which provides an open-source collection of web topics, but in a uniform distribution. 

We use a simple yet effective approach to produce a quasi-realistic topic distribution, based on topic modeling on the MS MARCO dataset \cite{bajaj2016ms}. More concretely, we use the Google Content Classification API (v2)\footnote{\url{https://cloud.google.com/natural-language/docs/classifying-text}} to classify a 300k subset of MS MARCO queries with scored labels from a set of 1091 topics/categories\footnote{\url{https://cloud.google.com/natural-language/docs/categories\#categories\_version_2}}. Since MS MARCO is based on real web queries, we expect it to provide a reliable distribution of topics in human queries.
To convert the result into a distribution that can be sampled with less sparsity, we fit a 2-label conditional probability distribution on the first two labels (highest scores) of our labeled samples. This results in $P(T_{1})$ (probability of the first topic), and $P(T_{2}|T_{1})$ (probability of the second topic conditioned on the first one). When generating the embedding data, we sample from these distributions for each inference and add $T_{1}$ and $T_{2}$ to the prompt. This enforces topic diversity across generated samples while adhering to a realistic distribution.   
\paragraph{Prompt Modification} We also add parameters to our generation prompts to improve (local) diversity as well as the quality or hardness of triplets. While the explicit topic sampling provides diverse topic seeds for prompts, the topic/category pool is quite universal and lacks regional variations. Considering that our final objective is to train a Dutch embedding model, we add a randomly activated flag ($p=.3$) to our prompt generation code which adds an instruction encouraging the model to generate the triplet in the Dutch/Flemish context, if possible (i.e. using regional events, entities, and references). As for the triplet quality, we add another control parameter that asks for the generated query to ``have minimum lexical overlap with the positive document'', in 50\% of the prompts. The final prompt templates can be found in Appendix D. 
\paragraph{Data Filtering}
Finally, we filter the generated triplets to remove samples with false positives and false or easy negatives. To do this, we use a high quality re-ranking model Qwen3-Reranker-4B \cite{qwen3embedding} to score the positive and negative documents, and apply a heuristic condition so that the difference between the positive and negative scores does not exceed a certain threshold\footnote{Based on our trials and considering the highly polarized re-ranker scores, we set $C=.96$ }, i.e.: $$ 0< S(p_{i}) - S(n_{i}) < C$$

\vspace{5pt}
We initially generate 500K triplets (200K for short-long, 140K for STS, 110K for long-short, 25K for short-short and 25K for long-long) following the ratios used by \citet{wang-etal-2024-improving-text}, which we then filter down to 350K triplets. For generation we use three OpenAI models (GPT-4.1, GPT-4.1-mini, and GPT-4.1-nano) and to optimize the quality/cost trade-off, we distribute the prompts between them based on an ad hoc 3-tier hardness measure (i.e. assigning more challenging prompts to more capable/costly models) that considers parameters like length, clarity, lexical overlap, and difficulty. Overall, the data generation process costs around 180 euro.

\subsection{Models}
This subsection describes the models used in our experiments for fine-tuning and benchmarking.

\subsubsection{Fine-tuned Models}
With the goal of developing compact and efficient Dutch models, we explore different initialization strategies for fine-tuning, using both supervised (models finetuned for embedding) and self-supervised models (PLM encoders).

\paragraph{Supervised models} For supervised models, we choose the E5 and mE5 suites (small, base and large) \cite{Wang2022TextEB,wang2024multilingual} based on their robust and high performance on the MTEB benchmark, and we investigate two initialization strategies to obtain weights tailored to Dutch: First, we leverage the Dutch capabilities of multilingual E5 and apply vocabulary trimming \cite{ushio-etal-2023-efficient} to reduce the vocabulary size from 250K to 50K tokens. This substantially reduces the model size by 66\% for small, 55\% for base, and 37\% for large.  We refer to these models as \textit{e5-trm}: e5-small-trm,  e5-base-trm,  e5-large-trm. Second, we use transtokeniser \cite{remy-delobelle2024transtokenization} to align and map the vocabulary of the English E5 suite (v2) with that of Bertje \cite{devries2019bertje}. We refer to these models as \textit{e5-t2t}: e5-small-v2-t2t, e5-base-v2-t2t, e5-large-v2-t2t. In both cases, the corresponding fine-tuned models are suffixed with \textit{-nl}. Both methods only modify the embedding matrix, not the main weights. However, the latter may add randomly initialized embeddings for untranslated tokens.

\paragraph{Self-supervised models} For self-supervised models, we use classic encoders trained with masked language modeling on Dutch data:  RobBERT-2023-base and  RobBERT-2023-large \citep{delobelle2023robbert2023conversion}. We select these models as they demonstrate strong performance on the DUMB benchmark. Previous work has shown that such models require a long stage of weak supervision to achieve strong performance \citep{wang-etal-2024-improving-text}, but we want to see how much improvement they gain from limited fine-tuning on labeled data, considering their optimized-for-Dutch attributes, including vocabulary. 
%However, previous work has shown that such models require weak supervision to achieve strong performance \citep{wang-etal-2024-improving-text}.  
%We fine-tune RobBERT-2023-base and RobBERT-2023-large, and t
The corresponding fine-tuned models are suffixed with \textit{-ft}.

\paragraph{Hyperparameters} For fine-tuning, we use in-batch negatives for all datasets except the classification subset of the synthetic data, where the relatively small number of labels can lead to false negatives. In addition, we include one hard negative alongside the in-batch negatives. Hard negatives are generated for the synthetic data (Section \ref{synth-data}) and mined for the existing retrieval data (Section \ref{existing-data} and Appendix D). We employ source-homogeneous batching, where all samples in a batch originate from the same dataset (one of the eight listed in Table\ref{tab:train_datasets_stat}), which reduces the risk of false in-batch negatives. We fine-tune the supervised models for one epoch and the self-supervised models for three, using the standard InfoNCE Loss. Training is carried out with a batch size of 1024, a learning rate of $2\times 10^{-6}$ for large models and $1\times 10^{-5}$ for the others, a warm-up ratio of 0.25, and a constant scheduler. For self-supervised models we set the learning rate and warm-up ratio to $2\times 10^{-5}$ and 0.1 respectively.

\subsubsection{Benchmarking Models}
We select a wide range of self-supervised and supervised models for our benchmarking experiments, focusing on models with fewer than 1B parameters. However, we also include Qwen3-Embedding-4B, one of the top-performing large models on multilingual MTEB. Our selection further covers a broad range of Dutch and multilingual models from the DUMB benchmark, as well as multilingual models incorporated in MTEB. We construct embeddings for self-supervised models by averaging token representations, while for other models we follow their specified requirements. 

Table~\ref{tab:models} (Appendix B) lists all models used in our experiments, along with their sources and citations.

\section{Results and Discussion}

\begin{table*}[h]
\centering
\resizebox{1\textwidth}{!}{ 
\begin{tabular}{lcccccccccc}
\hline
 & Prm & Cls & MLCls & PCls & Rrnk & Rtr & Clust. & STS & AvgD & AvgT \\
\midrule
Num. Datasets ($\rightarrow$) &  & 12 & 3 & 2 & 1 & 12 & 8 & 2 & 40 &  \\
\midrule

BERTje-base & 109M & 50.9 & 37.3 & 70.9 & 72.2 & 17.6 & \textbf{23.5} & 53.2 & \textbf{36.0} & \textbf{46.5} \\
Tik-to-Tok-base & 116M & 48.0 & 37.3 & 70.6 & 69.9 & 16.5 & 18.8 & 46.4 & 33.5  & 43.9 \\

RobBERT-v1-base & 117M & 48.2 & 33.9 & 70.4 & \textbf{72.8} & 16.4 & 20.7 & 52.0 & 34.0  & 44.9 \\
RobBERT-v2-base & 117M & 49.6 & 36.1 & 72.9 & 70.9 & 17.3 & 21.1 & 48.5 & 34.9  & 45.2 \\
RobBERT-2022-base & 119M & 49.5 & 36.8 & \textbf{73.4} & 67.1 & 13.7 & 22.7 & 51.7 & 34.2 & 45.0 \\
RobBERT-2023-base & 124M & 50.1 & 37.3 & 72.9 & 71.7 & \textbf{18.3} & 20.2 & 48.5 & 35.2 & 45.6 \\
mBERT-cased-base & 179M & 46.9 & 35.1 & 68.4 & 65.2 & 11.7 & 20.1 & 49.4 &  31.8 & 42.4 \\
mDeBERTa-v3-base & 184M  & 31.6 & 24.4 & 65.9 & 40.3 & 2.2 & 9.0 & 37.0 & 19.9  & 30.1 \\
XLM-R-base & 279M & 37.2 & 34.9 & 67.4 & 36.0 & 4.5 & 14.6 & 42.7 & 24.5  & 33.9 \\
Tik-to-Tok-large & 345M & 50.8 & 37.4 & 69.9 & 62.4 & 14.8 & \textbf{23.5} & \textbf{54.2} & 35.0  & 44.7 \\
RobBERT-2023-large & 355M & \textbf{52.0} & \textbf{37.6} & 68.9 & 57.1 & 11.9 & 23.3 & \textbf{54.2} & 34.2  & 43.6 \\
XLM-R-large & 561M & 39.1 & 35.5 & 64.3 & 39.4 & 8.8 & 13.5 & 39.9 & 25.9  & 34.4 \\
\hline
\midrule
static-similarity-mrl-multilingual-v1 & - & 49.4 & 33.2 & 63.6 & 80.5 & 37.9 & 15.1 & 64.5 & 40.1  & 49.2 \\
\textbf{e5-small-v2-t2t} & 33M &  53.7 & 38.5 & 74.5 & 85.9 & 45.0 & 24.1 & 74.3 & 46.9 & 56.6 \\
\textbf{e5-small-v2-t2t-nl}  & 33M  & 55.3 & 40.9 & 74.9 & 86.0 & 49.9 & 28.0 & 74.1 & 49.8 & 58.4 \\
\textbf{e5-small-trm} & 41M  & 56.3 & 43.5 & \textbf{76.5} & \textbf{87.3} & 53.1 & 28.2 & 74.2 & 51.4 & 59.9 \\
\textbf{e5-small-trm-nl} & 41M   & \textbf{58.2} & \textbf{44.7} & 76.0 & 87.1 & \textbf{56.0} & \textbf{32.2} & \textbf{74.6} & \textbf{53.8} & \textbf{61.3} \\
\midrule
granite-embedding-107m-multilingual & 107M & 53.9 &  41.8 &  70.1 & 84.7 & 50.2 &  29.8 & 68.4 & 49.4  & 57.0 \\
\textbf{e5-base-v2-t2t}  & 109M   & 54.4 & 40.3 & 73.3 & 85.6 & 46.2 & 25.5 & 73.2 & 47.8 & 56.9 \\
\textbf{e5-base-v2-t2t-nl}  & 109M  & 53.9 & 41.5 & 72.5 & 84.0 & 46.4 & 26.9 & 69.3 & 47.8 & 56.3  \\
multilingual-e5-small & 118M & 56.3 & 43.5 & 76.5 & 87.1 & 53.1 & 28.2 & 74.2 & 51.4  & 59.8 \\
paraphrase-multilingual-MiniLM-L12-v2 & 118M & 55.0 & 38.1 & 78.2 & 80.6 & 37.7 & 29.6 & 76.3 & 46.3  & 56.5 \\
\textbf{RobBERT-2023-base-ft} & 124M & 58.1 & 44.6 & 72.7 & 84.7 & 51.6 & 32.9 & 68.5 & 52.0 & 59.0 \\
\textbf{e5-base-trm} & 124M
  & 58.1 & 44.4 & 76.7 & 88.3 & 55.8 & 28.1 & 74.9 & 52.9 & 60.9 \\
\textbf{e5-base-trm-nl} & 124M   & \textbf{59.6} & \textbf{45.9} & 78.4 & 87.5 & 56.5 & \textbf{34.3} & 75.8 & \textbf{55.0} & \textbf{62.6} \\
potion-multilingual-128M & 128M & 51.8 & 40.0 & 60.4 & 80.3 & 35.7 & 26.1 & 62.0 & 42.6  & 50.9 \\
multilingual-e5-base & 278M & 58.2 & 44.4 & 76.7 & \textbf{88.4} & 55.8 & 27.7 & 74.9 &  52.8 & 60.9 \\
granite-embedding-278m-multilingual & 278M & 54.6 & 41.8 & 71.0 & 85.6 & 52.4 & 30.3 & 68.9 & 50.5 & 58.0 \\
paraphrase-multilingual-mpnet-base-v2 & 278M & 58.1 & 40.5 & \textbf{81.9} & 82.3 & 41.4 & 30.8 & 79.3 & 49.2  & 59.2 \\
Arctic-embed-m-v2.0 & 305M & 54.4  & 42.6 & 66.6  &  86.2 & 51.8  &  26.5 & 64.9  &  49.1 &  56.1 \\
gte-multilingual-base & 305M & 59.1  & 37.7 & 77.8  &  82.3 & \textbf{56.8}  &  31.3 & \textbf{78.6}  &  53.8 &  60.5 \\\midrule 
\textbf{e5-large-v2-t2t}   & 335M & 55.7 & 41.4 & 75.7 & 86.6 & 49.9 & 25.5 & 74.0 & 49.5 & 58.4 \\
\textbf{e5-large-v2-t2t-nl}   & 335M   & 57.3 & 42.4 & 76.9 & 86.9 & 50.8 & 27.7 & 74.1 & 51.7 & 59.4 \\
\textbf{RobBERT-2023-large-ft} & 355M  & 59.3 & 45.2 & 68.7 & 82.3 & 48.3 & 31.6 & 70.6 & 51.0 & 58.0 \\
\textbf{e5-large-trm} & 355M    & 60.2 & 45.4 & 80.3 & 90.3 & 59.0 & 28.7 & 78.8 & 55.1 & 63.3 \\
\textbf{e5-large-trm-nl} & 355M  & \textbf{62.2} & \textbf{48.0} & \textbf{81.4} & 87.2 & 58.2 & 35.6 & 78.2 & \textbf{57.0} & \textbf{64.4} \\
LaBSE & 470M & 54.6 & 41.4 & 76.0 & 80.5 & 35.5 & 25.2 & 69.0 & 44.5  & 54.6 \\
multilingual-e5-large & 560M & 60.2 & 45.4 & 80.3 & \textbf{90.3} & 59.1 & 29.5 & 78.8 & 55.3  & 63.4 \\
Arctic-embed-l-v2.0 & 568M & 59.3 & 45.2 & 74.2 & 88.2 & 59.0 & 29.8 & 71.7 & 54.3 & 61.1 \\
bge-m3 & 568M & 60.7 & 44.2 & 78.3 & 88.7 & \textbf{60.0} & 29.2 & 78.1 & 
55.4 & 63.1 \\
jina-embeddings-v3 & 572M & 61.7 & 38.9 & 76.8 & 78.5 & 59.1 & \textbf{38.9} & \textbf{84.8} & \textbf{57.0} & 62.7 \\
\hline\midrule 
KaLM-multilingual-mini-instruct-v1 & 594M & 59.4 & 45.9 & 76.4 & 85.5 & 54.2 & 39.0 & 73.4 & 54.9  & 62.0 \\
multilingual-e5-large-instruct & 560M & 63.6 & 47.4 & \textbf{80.6} & \textbf{88.2} & 61.4 & 46.0 & 81.3 & 60.6 & 66.9 \\
Qwen3-Embedding-0.6B & 596M & 60.5 & 46.4 & 73.6 & 85.1 & 57.1 & 38.7 & 76.8 &  56.1 & 62.6 \\
Qwen3-Embedding-4B & 4B & \textbf{68.9} & \textbf{50.5} & \textbf{80.6} & 87.1 & \textbf{64.2} & \textbf{47.1} & \textbf{85.8} &  \textbf{63.6} & \textbf{69.2} \\
\bottomrule
\end{tabular}
}
\caption{Results on MTEB-NL for self-supervised (top), supervised (middle), and supervised-instruct (bottom) models.  We split the middle section into three subsections: small models ($<$100M), base models ($<$305M), and large models ($>$305M).
 \textit{Prm} refers to the number of model parameters. Task categories are abbreviated as follows: classification (Cls), multilabel classification (MLClf), pair classification (PCls), reranking (Rrnk), retrieval (Rtr), clustering (Clust), and semantic textual similarity (STS). AvgD and AvgT refer to the averaged metrics per dataset and per task, respectively. We denote in bold the models introduced in this paper and the best results within each section. The fine-tuned models from this paper are marked with -nl and -ft suffixes.}
\label{table-mteb-nl-1}
\end{table*}

In this section, we present and discuss our results, which are shown in Table~\ref{table-mteb-nl-1}. Detailed results per dataset are provided in Table~\ref{tab:all_datasets_with_average} in Appendix~B.

\paragraph{Self-supervised models} As expected, self-supervised models (the top section of Table~\ref{table-mteb-nl-1}) are outperformed by supervised models. However, in classification and pair-classification tasks they achieve competitive results compared to some supervised models. The widest gap is observed in retrieval, as masked language modeling does not explicitly capture this task. In addition, multilingual models (mBERT-cased-base, mDeBERTa-v3-base, XLM-R-base, and XLM-R-large) are the weakest performers among the self-supervised models. This clearly highlights the benefits of adapting encoders to Dutch. BERTje-base demonstrates the best average performance, although it only shows superior performance in a single task (clustering).

\paragraph{Supervised models} The best results are obtained from the supervised-instruct models (the bottom section of Table~\ref{table-mteb-nl-1}: multilingual-e5-large-instruct and Qwen3-Embedding-4B), which outperform all other models by a substantial margin. Moreover, larger models consistently demonstrate stronger performance. The fine-tuned models introduced in this paper dominate the category of non-instruct models, while being more efficient in terms of parameters. e5-small-trm-nl consistently outperforms base-sized models and even surpasses some large models (LaBSE, Arctic-embed-l-v2.0). e5-base-trm-nl achieves performance comparable to jina-embeddings-v3 and Qwen3-Embedding-0.6B, despite being four times smaller. e5-large-trm-nl is the best non-instuct model and even outperforms some of the instruct models, such as Qwen3-Embedding-0.6B and KaLM-multilingual-mini-instruct-v1. However, it lags behind the two best instruct models.

\paragraph{Initialization strategies} The -trm models (mE5 with trimmed vocabulary) demonstrate better performance than the -t2t models (E5 with translated vocabulary) of the same size. In addition, the -trm models retain the same quality as their multilingual counterparts while being considerably smaller. This is also reflected in the fine-tuning results, where the fine-tuned -trm models are superior. RobBERT-2023-base-ft and RobBERT-2023-large-ft lag behind the top-performing models. However, they achieve surprisingly competitive results even without weak supervision.  We assume that these models could benefit greatly from this step.

\section{Conclusion and Future Work}

In this work, we addressed the lack of Dutch embedding resources by introducing the MTEB-NL benchmark, training datasets, and models tailored to the Dutch language. 
We released MTEB-NL, a comprehensive benchmark that combines existing and newly created datasets across a wide range of tasks, resulting in 40 datasets. 

The current benchmark still relies on machine- and human-translated resources, which may overlook linguistic nuances and cultural context specific to Dutch. Future work should reduce reliance on translated data and focus on creating resources more closely tailored to the Dutch context. In addition, the current version of the benchmark is not balanced across tasks, with some tasks represented by only one or two datasets. Future work should address this imbalance.

To support model training, we compiled a Dutch retrieval dataset and augmented it with synthetic data to ensure broader task coverage. Building on this, we introduced the E5-NL model suite, compact yet efficient embedding models derived from E5. Our models achieve superior performance among non-instruct models and even outperform some instruct models. Hence, the next logical step is the development of instruct models for Dutch, which should be a focus of future work. We demonstrated that Dutch self-supervised encoders can achieve competitive results but still lag behind supervised models. They can be good candidates to fill this niche if weak supervision is involved, as was shown in previous work \citep{wang-etal-2024-improving-text}. An alternative approach is to adapt Dutch generative models such as ChocoLLama \citep{meeus2024chocollama}. However, the latter approach requires larger computational resources for experiments, compared to encoders.

All resources are made publicly available through the Hugging Face Hub and the MTEB package, with the aim of fostering further research and development of Dutch embeddings.

\section{Acknowledgements}
This research received funding from the Flemish
Government under the “Onderzoeksprogramma Artificiële Intelligentie (AI) Vlaanderen” programme. In addition, we acknowledge the use of ChatGPT for
assisting with error checking and proofreading of
this paper.

\section{Limitations}

\paragraph{Native Dutch Resources} While MTEB-NL provides a benchmark for evaluating embedding models in Dutch, it still relies partly on machine- and human-translated datasets. This reliance limits its ability to fully capture the linguistic nuances and cultural context of Dutch. As mentioned earlier, future versions of MTEB-NL should aim to reduce the proportion of translated content and move towards more authentic native data.

\paragraph{Benchmark Validity Over Time} The first version of MTEB has become a standard benchmark for evaluating embedding models, attracting numerous evaluations over time. Such extensive use introduces the risk of overfitting, as researchers may inadvertently optimize models for strong performance on MTEB rather than for generalization. Moreover, many submissions have relied on training sets drawn from the same datasets, meaning that comparisons are no longer strictly zero-shot. These issues apply equally to MTEB-NL, and we therefore emphasize the importance of using it in the zero-shot setting. In addition, the rapid pace of advances in embedding models and evolving evaluation needs may outgrow the benchmark, making it less representative and less relevant over time.

\nocite{*}
\section{Bibliographical References}\label{sec:reference}

\bibliographystyle{lrec2026-natbib}
\bibliography{lrec2026-example}

\section{Language Resource References}
\label{lr:ref}
\bibliographystylelanguageresource{lrec2026-natbib}
\bibliographylanguageresource{languageresource}

\section{Appendix A. MTEB-NL Datasets}
\begin{figure*}[h]
    \centering
    \includegraphics[width=1\linewidth]{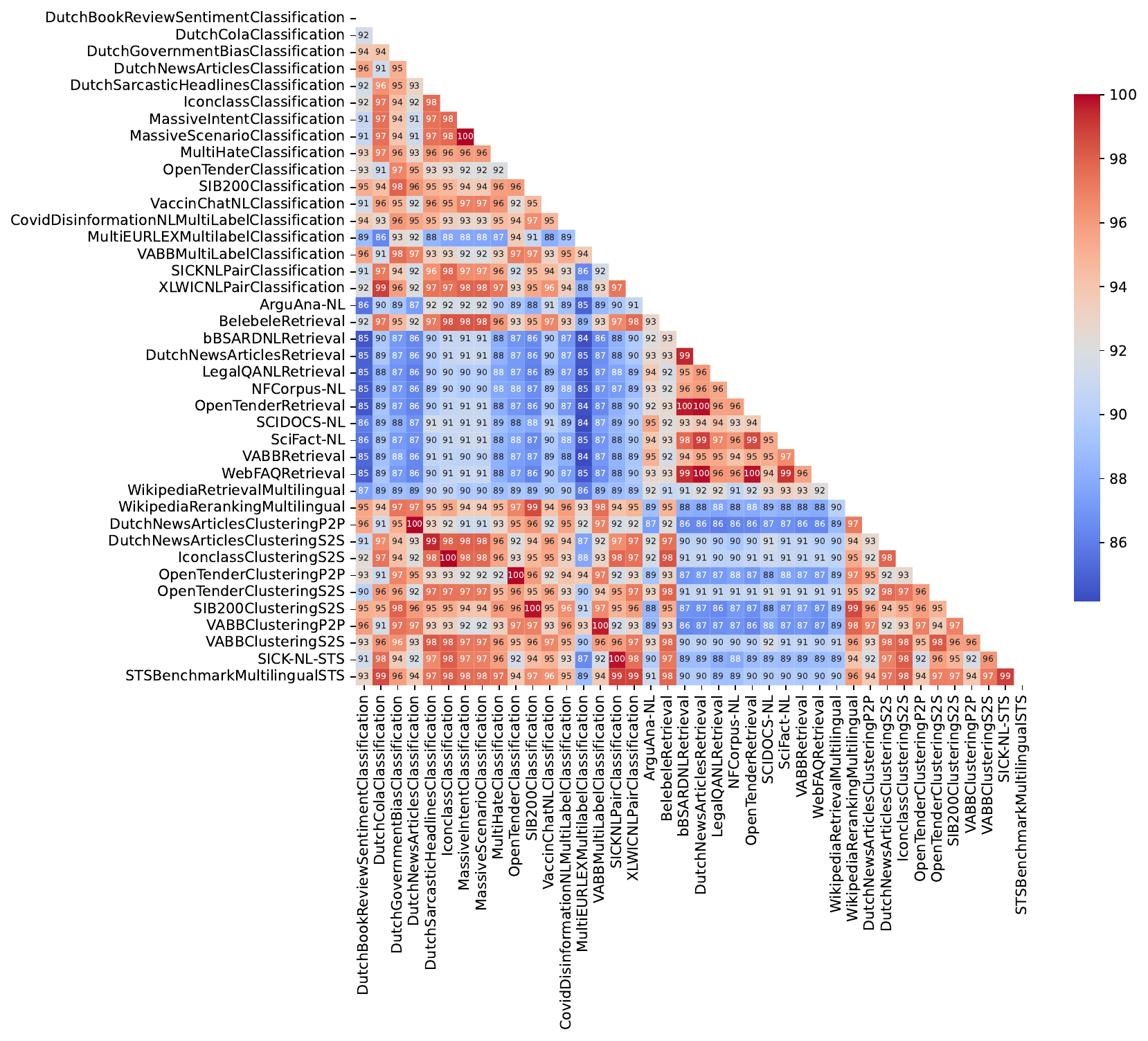} % or plot.png
    \caption{Similarity within MTEB-NL datasets. We use the \textit{e5-multilingual-base} model to embed 100 documents from each dataset. Then, we compute cosine similarities between the averaged embeddings for the visualization.}
    \label{fig:similarity}
\end{figure*}

For our experiments, we rely on the MTEB package, adopting the setup introduced in MMTEB. Accordingly, the task definitions and settings are fully aligned with MMTEB. Table \ref{tab:dutch_datasets} lists all datasets used, including their citations and corresponding URLs on Hugging Face. Table \ref{tab:dutch_datasets_stat} presents statistics on these datasets. Figure \ref{fig:similarity} shows the similarity of documents in our benchmarks. Below, we describe the datasets and preprocessing steps used in MTEB-NL. All datasets were converted to the MTEB format, except for those already included in MTEB.

\begin{table*}[htbp]
\centering
% \hspace*{cm}
\small
\resizebox{1\textwidth}{!}{ 
\begin{tabular}{@{}p{.45\textwidth} p{.35\textwidth} p{.20\textwidth} p{.45\textwidth}@{}}
\toprule
\textbf{Dataset} & \textbf{Source} & \textbf{License} & \textbf{Hugging Face URL} \\ \midrule
DutchBookReviewSentimentClassification & \citet{van2019merits} & \textsc{CC-BY-NC-SA-4.0} & \href{https://huggingface.co/datasets/mteb/DutchBookReviewSentimentClassification}{\nolinkurl{mteb/DutchBookReviewSentiment...}} \\
DutchColaClassification & \citetlanguageresource{gronlp_2024} & unknown & \href{https://huggingface.co/datasets/GroNLP/dutch-cola}{\nolinkurl{GroNLP/dutch-cola}} \\
DutchGovernmentBiasClassification & \citet{de2025detecting} & \textsc{CC-BY-NC-SA-4.0} & \href{https://huggingface.co/datasets/milenamileentje/Dutch-Government-Data-for-Bias-detection}{\nolinkurl{milenamileentje/Dutch-Government..}} \\
DutchNewsArticlesClassification & \citet{max_scheijen_2022} & \textsc{CC-BY-NC-SA-4.0} & \href{https://huggingface.co/datasets/clips/mteb-nl-news-articles-cls}{\nolinkurl{clips/mteb-nl-news-articles-cls}} \\
DutchSarcasticHeadlinesClassification & \citet{tuin_dutch_news_headlines_2020} & \textsc{CC0} & \href{https://huggingface.co/datasets/clips/mteb-nl-sarcastic-headlines}{\nolinkurl{clips/mteb-nl-sarcastic-headlines}} \\
IconclassClassification & \citet{banar2023transfer} & \textsc{CC-BY-NC-SA-4.0} & \href{https://huggingface.co/datasets/clips/mteb-nl-iconclass-cls}{\nolinkurl{clips/mteb-nl-iconclass-cls}} \\
MassiveIntentClassification & \citet{fitzgerald2023massive} & \textsc{Apache 2.0} & \href{https://huggingface.co/datasets/mteb/amazon_massive_intent}{\nolinkurl{mteb/amazon_massive_intent}} \\
MassiveScenarioClassification & \citet{fitzgerald2023massive} & \textsc{Apache 2.0} & \href{https://huggingface.co/datasets/mteb/amazon_massive_scenario}{\nolinkurl{mteb/amazon_massive_scenario}} \\
MultiHateClassification & \citet{rottger-etal-2022-multilingual} & \textsc{CC-BY-4.0} & \href{https://huggingface.co/datasets/mteb/multi-hatecheck}{\nolinkurl{mteb/multi-hatecheck}} \\
OpenTenderClassification & Introduced by our paper & \textsc{CC-BY-NC-SA-4.0} & \href{https://huggingface.co/datasets/clips/mteb-nl-opentender-cls}{\nolinkurl{clips/mteb-nl-opentender-cls}} \\
SIB200Classification & \citet{adelani-etal-2024-sib} & \textsc{CC-BY-SA-4.0} & \href{https://huggingface.co/datasets/mteb/sib200}{\nolinkurl{mteb/sib200}} \\
VaccinChatNLClassification & \citet{buhmann-etal-2022-domain} & \textsc{CC-BY-4.0} & \href{https://huggingface.co/datasets/clips/VaccinChatNL}{\nolinkurl{clips/VaccinChatNL}} \\
\midrule 
CovidDisinformationNLMultiLabelClassification & \citet{alam-etal-2021-fighting-covid} & \textsc{CC-BY-4.0} & \href{https://huggingface.co/datasets/clips/mteb-nl-COVID-19-disinformation}{\nolinkurl{clips/mteb-nl-COVID-19-disinf...}} \\
MultiEURLEXMultilabelClassification & \citet{chalkidis-etal-2021-multieurlex} & \textsc{CC-BY-SA-4.0} & \href{https://huggingface.co/datasets/mteb/eurlex-multilingual}{\nolinkurl{mteb/eurlex-multilingual}} \\
VABBMultiLabelClassification & \citetlanguageresource{aspeslagh2024vabbshw_blx} & \textsc{CC-BY-4.0} & \href{https://huggingface.co/datasets/clips/mteb-nl-vabb}{\nolinkurl{clips/mteb-nl-vabb}} \\
 \midrule 
SICKNLPairClassification & \citet{wijnholds-moortgat-2021-sick} & \textsc{MIT} & \href{https://huggingface.co/datasets/clips/mteb-nl-sick}{\nolinkurl{clips/mteb-nl-sick}} \\
XLWICNLPairClassification & \citet{raganato2020xl} & \textsc{CC-BY-NC-4.0} & \href{https://huggingface.co/datasets/pasinit/xlwic}{\nolinkurl{pasinit/xlwic}} \\
 \midrule 
ArguAna-NL & \citet{lotfi2025beir} & \textsc{CC-BY-SA-4.0} & \href{https://huggingface.co/datasets/clips/beir-nl-arguana}{\nolinkurl{clips/beir-nl-arguana}} \\
BelebeleRetrieval & \citet{bandarkar-etal-2024-belebele} & \textsc{CC-BY-SA-4.0} & \href{https://huggingface.co/datasets/facebook/belebele}{\nolinkurl{facebook/belebele}} \\
bBSARDNLRetrieval & \citet{lotfi-etal-2025-bilingual} & \textsc{CC-BY-NC-SA-4.0} & \href{https://huggingface.co/datasets/clips/bBSARD}{\nolinkurl{clips/bBSARD}} \\
DutchNewsArticlesRetrieval & \citet{max_scheijen_2022} & \textsc{CC-BY-NC-SA-4.0} & \href{https://huggingface.co/datasets/clips/mteb-nl-news-articles-ret}{\nolinkurl{clips/mteb-nl-news-articles-ret}} \\
LegalQANLRetrieval & \citet{redelaar-etal-2024-attributed} & unknown & \href{https://huggingface.co/datasets/clips/mteb-nl-legalqa}{\nolinkurl{clips/mteb-nl-legalqa}} \\
NFCorpus-NL & \citet{lotfi2025beir} & \textsc{CC-BY-SA-4.0} & \href{https://huggingface.co/datasets/clips/beir-nl-nfcorpus}{\nolinkurl{clips/beir-nl-nfcorpus}} \\
OpenTenderRetrieval & Introduced by our paper & \textsc{CC-BY-NC-SA-4.0} & \href{https://huggingface.co/datasets/clips/mteb-nl-opentender-ret}{\nolinkurl{clips/mteb-nl-opentender-ret}} \\
SCIDOCS-NL & \citet{lotfi2025beir} & \textsc{CC-BY-SA-4.0} & \href{https://huggingface.co/datasets/clips/beir-nl-scidocs}{\nolinkurl{clips/beir-nl-scidocs}} \\
SciFact-NL & \citet{lotfi2025beir} & \textsc{CC-BY-SA-4.0} & \href{https://huggingface.co/datasets/clips/mteb-nl-nq}{\nolinkurl{clips/mteb-nl-nq}} \\
VABBRetrieval & \citetlanguageresource{aspeslagh2024vabbshw_blx} & \textsc{CC-BY-4.0} & \href{https://huggingface.co/datasets/clips/mteb-nl-vabb-ret}{\nolinkurl{clips/mteb-nl-vabb-ret}} \\
WebFAQRetrieval & \citet{dinzinger2025webfaq} & \textsc{CC-BY-4.0} & \href{https://huggingface.co/datasets/PaDaS-Lab/webfaq}{\nolinkurl{PaDaS-Lab/webfaq}} \\
WikipediaRetrievalMultilingual & \citet{wikidump} & \textsc{CC-BY-SA-3.0} & \href{https://huggingface.co/datasets/ellamind/wikipedia-2023-11-retrieval-multilingual-queries}{\nolinkurl{ellamind/wikipedia-2023-11-ret..}} \\\midrule
WikipediaRerankingMultilingual & \citet{wikidump} & \textsc{CC-BY-SA-3.0} & \href{https://huggingface.co/datasets/ellamind/wikipedia-2023-11-reranking-multilingual}{\nolinkurl{ellamind/wikipedia-2023-11-rerank..}} \\\midrule
DutchNewsArticlesClusteringP2P & \citet{max_scheijen_2022} & \textsc{CC-BY-NC-SA-4.0} & \href{https://huggingface.co/datasets/clips/mteb-nl-news-articles-cls}{\nolinkurl{clips/mteb-nl-news-articles-cls}} \\
DutchNewsArticlesClusteringS2S & \citet{max_scheijen_2022} & \textsc{CC-BY-NC-SA-4.0} & \href{https://huggingface.co/datasets/clips/mteb-nl-news-articles-cls}{\nolinkurl{clips/mteb-nl-news-articles-cls}} \\
IconclassClusteringS2S & \citet{banar2023transfer} & \textsc{CC-BY-NC-SA-4.0} & \href{https://huggingface.co/datasets/clips/mteb-nl-iconclass-cls}{\nolinkurl{clips/mteb-nl-iconclass-cls}} \\
OpenTenderClusteringP2P & Introduced by our paper & \textsc{CC-BY-NC-SA-4.0} & \href{https://huggingface.co/datasets/clips/mteb-nl-opentender-cls}{\nolinkurl{clips/mteb-nl-opentender-cls}} \\
OpenTenderClusteringS2S & Introduced by our paper & \textsc{CC-BY-NC-SA-4.0} & \href{https://huggingface.co/datasets/clips/mteb-nl-opentender-cls}{\nolinkurl{clips/mteb-nl-opentender-cls}} \\
SIB200ClusteringS2S & \citet{adelani-etal-2024-sib} & \textsc{CC-BY-SA-4.0} & \href{https://huggingface.co/datasets/mteb/sib200}{\nolinkurl{mteb/sib200}} \\
VABBClusteringP2P & \citetlanguageresource{aspeslagh2024vabbshw_blx} & \textsc{CC-BY-4.0} & \href{https://huggingface.co/datasets/clips/mteb-nl-vabb-cls}{\nolinkurl{clips/mteb-nl-vabb-cls}} \\
VABBClusteringS2S & \citetlanguageresource{aspeslagh2024vabbshw_blx} & \textsc{CC-BY-4.0} & \href{https://huggingface.co/datasets/clips/mteb-nl-vabb-cls}{\nolinkurl{clips/mteb-nl-vabb-cls}} \\

\midrule 
SICK-NL-STS & \citet{wijnholds-moortgat-2021-sick} & \textsc{MIT} & \href{https://huggingface.co/datasets/clips/mteb-nl-sick}{\nolinkurl{clips/mteb-nl-sick}} \\
STSBenchmarkMultilingualSTS & \citet{huggingface:dataset:stsb_multi_mt} & unknown & \href{https://huggingface.co/datasets/mteb/stsb_multi_mt}{\nolinkurl{mteb/stsb_multi_mt}} \\

\bottomrule 
\end{tabular}
}
\caption{Overview of Dutch datasets and their Hugging Face entries with licenses and citations.}
\label{tab:dutch_datasets}
\end{table*}

\subsection{Classification}

A subsample of a training set (ranging from 8 to 128 samples) and a test set are embedded using the provided model. If only a test set is available, a portion of it is used as a training set. The training set embeddings are used to fit a logistic regression classifier with a maximum of 100 iterations (10 times repeated), which is then evaluated on the test set. We report the F1-macro score as the main evaluation metric.

\paragraph{DutchBookReviewSentimentClassification} A collection of book reviews retrieved from the \textit{Hebban} website, which are annotated with binary sentiment labels (1 indicating positive, 0 indicating negative). The binary labels are obtained by converting the original book rating scores, ranging from 1 to 5. This dataset does not involve any additional preprocessing, as it is already included in MTEB.

\paragraph{DutchColaClassification} This dataset consists of sentences in standard Dutch that are retrieved from the \textit{Syntax of Dutch} book series. The sentences are marked as acceptable (1) or not acceptable (0). No additional preprocessing steps were applied to this dataset.

\paragraph{DutchGovernmentBiasClassification} This dataset is obtained from the Dutch House of Representatives after 2010, which are annotated with a binary label for bias (1 indicating bias, 0 indicating non-bias). The dataset is used in its original form, without further preprocessing.

\paragraph{DutchNewsArticlesClassification} This dataset contains Dutch news articles obtained from \textit{Nederlandse Oproep Stichting (NOS)}. We use the articles belonging to the eight most frequent categories, sampled equally across categories. All duplicate items were removed.

\paragraph{DutchSarcasticHeadlinesClassification} This dataset contains news headlines from a satirical news website (\textit{Speld.nl}) and a regular news website that are annotated with binary sarcasm labels (1 indicating sarcasm, 0 indicating non-sarcasm). All headlines from \textit{Speld.nl} are annotated as sarcastic, whereas all headlines from \textit{nu.nl} are not. The dataset is sourced from Kaggle. Stratified sampling is applied based on the features \texttt{is\_sarcastic}, \texttt{is \_binnenland}, \texttt{is\_buitenland}, and \texttt{is\_politiek}. Duplicate entries were removed.

\paragraph{IconclassClassification}  Iconclass is a hierarchical classification system designed for categorizing the subjects and content of artworks. Each artwork can be annotated with one or multiple iconclass codes, each describing a specific concept in a painting, sourced from the \textit{Netherlands Institute for Art History}, which contains annotations for artwork titles. Despite the hierarchical structure and multi-label nature of Iconclass codes, we restrict our setup to the first layer in a multi-class classification setting, due to the complexity of the task and the rarity of items with more than two labels. This results in nine main Iconclass categories (e.g., Religion and Magic, Nature, Bible, etc.).

\paragraph{MassiveIntentClassification} This dataset is part of the human-translated and localized version of the SLURP NLU dataset \citep{bastianelli-etal-2020-slurp}. All instances, which are transcribed Virtual Assistants (VA) voice commands, are annotated with one of 60 different intents, i.e. actions that the VA has to perform. The dataset was used as provided, without additional preprocessing, as it is already incorporated in MTEB.

\paragraph{MassiveScenarioClassification} Similarly to MassiveIntentClassification, this dataset comprises virtual assistant commands that were human-translated from English to Dutch. These instances are annotated with one of 18 distinct scenarios. No preprocessing applied, as the dataset is already in MTEB.

\paragraph{MultiHateClassification} This dataset contains potential hate-speech test cases that were human-translated from English to Dutch. The instances are assigned a binary label (1 indicating hate-speech, 0 indicating non-hate-speech). The dataset was employed as provided, without further preprocessing, since it is already integrated in MTEB.

\paragraph{OpenTenderClassification} This dataset contains Belgian and Dutch tender calls from \textit{OpenTender}\footnote{\url{https://opentender.eu/}} in Dutch. Each tender, consisting of a title and description, is labeled with one of the 30  most frequent CPV-codes\footnote{For an extensive overview of all possible CPV-codes and their corresponding descriptions, consult \url{https://www.publictendering.com/list-of-the-cpv-codes/}.}, which denote the topic of the tender in question, from the first layer in the hierarchical structure.

\paragraph{SIB200Classification} This dataset contains sentences from the Flores-200 \cite{nllb2022} dataset that were translated to Dutch by professional translators. Each sentence, obtained from a web article, is assigned one of 7 possible topic labels from WikiNews, WikiJunior and WikiVoyage. The dataset was used as provided, without additional preprocessing, since it is already included in MTEB.

\paragraph{VaccinChatNLClassification} This is a Dutch COVID FAQ answer dataset that consists of questions asked by users of a COVID chatbot. All questions are annotated with one of the 181 possible intents (or answer classes). The test split covers 161 of these intents. This dataset does not involve any additional preprocessing steps.

\subsection{Multilabel Classification}
The MTEB framework downsamples the training sets for 10 experiments, restricting them to a fixed number of instances per unique label (ranging from 8 to 128 in our case). A K-Nearest Neighbors classifier is then trained on each set and evaluated on the same test set. For evaluation, we use F1-macro as the main metric.

\paragraph{CovidDisinformationNLMultiLabelClassification} This dataset contains Dutch tweets that were annotated for fine-grained disinformation analysis, with 7 possible labels. We filter the data using the variable indicating whether a tweet is understandable and use the remaining six labels for multi-label classification.

\paragraph{MultiEURLEXMultilabelClassification} This dataset contains the Dutch subset of European parliamentary documents from the MultiEURLEX dataset. All documents are annotated with one or multiple EUROVOC concepts related to the legal domain. No additional preprocessing was required, as this dataset is already integrated into MTEB.

\paragraph{VABBMultiLabelClassification} We use the Dutch subset of the VABB dataset (\textit{Flemish Academic Bibliography for the Social Sciences and Humanities}) from \citetlanguageresource{aspeslagh2024vabbshw_blx}. Each abstract, together with its corresponding title, is annotated with one or more academic disciplines, with 19 possible labels in total. The train, development, and test splits are stratified.

\begin{table*}[htbp]
\centering
\small
\resizebox{1\textwidth}{!}{ 
\begin{tabular}{@{}p{.25\textwidth} p{.20\textwidth} p{.40\textwidth} p{.11\textwidth} p{.11\textwidth}@{}}
\toprule
\textbf{Type} & \textbf{Domain} & \textbf{Dataset} & \textbf{n\_sources} & \textbf{n\_targets} \\ \midrule
Classification & Reviews & DutchBookReviewSentimentClassification & 2224 & 2 \\
Classification & Linguistics & DutchColaClassification & 2400 & 2 \\
Classification & Government & DutchGovernmentBiasClassification & 782 & 2 \\
Classification & News & DutchNewsArticlesClassification & 1200 & 8 \\
Classification & News & DutchSarcasticHeadlinesClassification & 1326 & 2 \\
Classification & Arts & IconclassClassification & 202 & 9 \\
Classification & Spoken & MassiveIntentClassification & 2974 & 60 \\
Classification & Spoken & MassiveScenarioClassification & 2974 & 18 \\
Classification & Hate speech  & MultiHateClassification & 1000 & 2 \\
Classification & Government & OpenTenderClassification & 4500 & 30 \\
Classification & News & SIB200Classification & 204 & 7 \\
Classification & Medical & VaccinChatNLClassification & 1170 & 161 \\ 
\midrule
MultiLabel Classification & Social Media & CovidDisinformationNLMultiLabelClassification & 252 & 7 \\
MultiLabel Classification & Legal & MultiEURLEXMultilabelClassification & 5000 & 21 \\
MultiLabel Classification & Scientific & VABBMultiLabelClassification & 3245 & 19 \\
\midrule
Pair Classification & Web & SICKNLPairClassification & 2116 & 2 \\
Pair Classification & Linguistics & XLWICNLPairClassification & 1004 & 2 \\
\midrule
Retrieval & Medical & ArguAna-NL & 8674 & 1406 \\
Retrieval & Web & BelebeleRetrieval & 488 & 900 \\
Retrieval & Legal & bBSARDNLRetrieval & 22417 & 222 \\
Retrieval & News & DutchNewsArticlesRetrieval & 255524 & 1000 \\
Retrieval & Legal & LegalQANLRetrieval & 30803 & 102 \\
Retrieval & Medical  & NFCorpus-NL & 3633 & 323 \\
Retrieval & Government & OpenTenderRetrieval & 137633 & 1000 \\
Retrieval & Scientific & SCIDOCS-NL & 25657 & 1000 \\
Retrieval & Scientific & SciFact-NL & 5183 & 300 \\
Retrieval & Scientific & VABBRetrieval & 9254 & 1000 \\
Retrieval & Web & WebFAQRetrieval & 370662 & 10000 \\
Retrieval & Wikipedia & WikipediaRetrievalMultilingual & 13,500 & 1500 \\
\midrule
Reranking & Wikipedia & WikipediaRerankingMultilingual & 1500 & 1500/12000 \\
\midrule
Clustering & News & DutchNewsArticlesClusteringP2P & 1200 & 8 \\
Clustering & News & DutchNewsArticlesClusteringS2S & 1200 & 8 \\
Clustering & Arts & IconclassClusteringS2S & 202 & 9 \\
Clustering & Government & OpenTenderClusteringP2P & 4500 & 30 \\
Clustering & Government & OpenTenderClusteringS2S & 4500 & 30 \\
Clustering & News & SIB200ClusteringS2S & 1004 & 7 \\
Clustering & Scientific & VABBClusteringP2P & 195 & 13 \\
Clustering & Scientific & VABBClusteringS2S & 195 & 13 \\
\midrule
STS & Web & SICK-NL-STS & 4096 & [1, 5] \\
STS & Miscellaneous & STSBenchmarkMultilingualSTS & 1379 & [0, 5] \\
\bottomrule 
\end{tabular}
}
\caption{Overview of the Dutch datasets included in MTEB-NL.
\textit{Domain} denotes the subject area/source. 
\textit{n\_sources} denotes corpus size(s) in retrieval tasks and test set size(s) in other tasks. 
\textit{n\_targets} refers to the number of labels in classification and clustering tasks, 
the number of queries in retrieval tasks, 
the number of positive/negative examples in reranking tasks, 
and the range of scores in STS tasks.}
\label{tab:dutch_datasets_stat}
\end{table*}

\subsection{Pair Classification}
The pair classification task involves assigning a binary label to two text inputs, embedded by a model. Their similarity is measured using various distance metrics (cosine similarity, dot product, Euclidean distance, Manhattan distance, or a model-specific similarity function). The primary evaluation metric is the average precision score computed at the optimal binary threshold.

\paragraph{SICKNLPairClassification}This dataset is the Dutch machine-translated and human post-edited version of the original SICK dataset \cite{marelli2014sick}, which was constructed from image and video caption descriptions. The dataset is annotated with one of three possible labels: entailment, contradiction, or neutral. Following the experimental setup of MTEB, we exclude the neutral label, resulting in a binary classification task.

\paragraph{XLWICNLPairClassification} This dataset is the Dutch subset of the XL-WiC benchmark, based on Multilingual WordNet and Wiktionary, and designed for the Word-in-Context task. In this task, the same target word appears in two different contexts, and the objective is to determine whether the word has the same meaning in both. The labels are binary: True (same meaning) or False (different meaning). The dataset was used as provided, without additional preprocessing.

\subsection{Retrieval}
The retrieval task aims to identify the documents relevant to a given query. Both queries and corpus documents are embedded using the provided model, and similarity scores are computed with cosine similarity. Retrieval performance is primarily evaluated using nDCG@10.

\paragraph{ArguAna-NL} ArguAna-NL is the Dutch machine-translated version of the original ArguAna dataset \cite{wachsmuth-etal-2018-retrieval}, an argument retrieval dataset from BEIR. It consists of claim–document pairs, where the task is to retrieve counterarguments for a given claim. The dataset is part of BEIR-NL and has already been adapted to the MTEB format.

\paragraph{BelebeleRetrieval} Belebele is a machine reading comprehension dataset consisting of passages paired with multiple-choice questions, where the task is to select the correct answer. In MTEB, this dataset is converted into a retrieval setup. We use the Dutch portion of the dataset, which was employed as provided, without additional preprocessing.

\paragraph{bBSARDNLRetrieval} bBSARD is a bilingual extension of the Belgian Statutory Article Retrieval Dataset (BSARD; \citealt{louis-spanakis-2022-statutory}), containing parallel statutory articles in Dutch and French. We use the Dutch subset of this dataset. It contains legislative articles sourced from official Belgian resources, along with queries machine-translated from French. The dataset is designed for article retrieval tasks in a legal setting, where the goal is to retrieve relevant statutory articles in response to legal queries. The dataset was used as provided, without additional preprocessing.

\paragraph{DutchNewsArticlesRetrieval} We use the same data source as in DutchNewsArticlesClassification to create a retrieval dataset. The task involves matching a subsample of titles with their corresponding articles.

\paragraph{LegalQANLRetrieval} This dataset consists of Dutch legislative articles, with queries formulated based on subordinating conjunctions. In this sense, the dataset is similar to bBSARDNLRetrieval, but it covers legislation from the Netherlands. The dataset was used without additional preprocessing.

\paragraph{NFCorpus-NL} NFCorpus-NL is the machine-translated Dutch version  of the NFCorpus dataset \cite{boteva2016full} from BEIR, created as part of the BEIR-NL benchmark. It consists of consumer health questions, medical and health-related documents, and graded relevance judgments. The dataset is included in BEIR-NL and has already been adapted to the MTEB format.
\paragraph{OpenTenderRetrieval} We use the same data source as in OpenTenderClassification to construct a retrieval dataset. The task consists of matching a subsample of titles with the corresponding tender descriptions.

\paragraph{SCIDOCS-NL} SCIDOCS-NL is the machine-translated Dutch adaptation of the SCIDOCS dataset \cite{cohan-etal-2020-specter} from BEIR, released as part of BEIR-NL. The task is to retrieve documents that are cited, or should be cited, for a paper given its title as input. SCIDOCS-NL is already included in BEIR-NL and has been adapted to the MTEB format.

\paragraph{SciFact-NL} SciFact-NL is the Dutch version of the SciFact dataset \cite{wadden-etal-2020-fact} from BEIR, machine-translated for the BEIR-NL benchmark. The original SciFact dataset contains scientific claims as queries and abstracts from scientific papers as documents, with relevance labels indicating whether a document supports or refutes a given claim. SciFact-NL is included in BEIR-NL and has already been adapted to the MTEB format.
\paragraph{VABBRetrieval} We use the same data source as in VABBRetrievalMultiLabelClassification to build a retrieval dataset. The task involves matching a subsample of titles with their corresponding abstracts.
\paragraph{WebFAQRetrieval} WebFAQRetrieval is a broad-coverage corpus of natural question–answer pairs in multiple languages, collected from FAQ pages on the web. We use the Dutch subset of this dataset. WebFAQRetrieval is already included in MTEB and was used as provided, without additional preprocessing.
\paragraph{WikipediaRetrievalMultilingual} This dataset is a multilingual retrieval dataset constructed from queries generated by a multilingual LLM grounded in Wikipedia articles. It was designed to resemble SQuAD \citep{rajpurkar-etal-2016-squad}. The dataset was used as provided, without further preprocessing, as it is already integrated into MTEB.

\subsection{Reranking}
The inputs consist of a query along with relevant and irrelevant documents. The goal is to rank the documents by their relevance to the query. The model embeds the documents, which are then compared to the query using cosine similarity. The primary metric is MAP. 

\paragraph{WikipediaRerankingMultilingual} This dataset is constructed similarly to \textit{WikipediaRetrievalMultilingual}. We use the Dutch portion, where each instance consists of a query paired with candidate passages annotated as either positive (relevant) or negative (irrelevant). The dataset was used as provided, without further preprocessing, as it is already integrated into MTEB.

\subsection{Clustering}
Clustering datasets consist of a collection of documents together with their associated labels. As our clustering datasets are small, we embed all documents (as opposed to MMTEB, which uses a subsample from the original set). The embeddings are then clustered using K-means, and performance is assessed by comparing the resulting clusters with the ground-truth labels.

The following datasets correspond to their classification counterparts: DutchNewsArticlesClusteringP2P, DutchNewsArticlesClusteringS2S, IconclassClusteringS2S, OpenTenderClusteringP2P, OpenTenderClusteringS2S, and SIB200ClusteringS2S. VABBClusteringP2P and VABBClusteringS2S are derived from the main dataset, sampled with one label per instance and an equal label distribution. Here, S2S denotes setups where only the title is used, while P2P refers to cases where both the title and passage are included.

\subsection{STS}
STS tasks consist of sentence pairs, with the objective of assessing their degree of similarity. Labels are continuous scores, where higher values correspond to greater similarity. Each sentence is embedded using the model, and pairwise similarity is calculated with various distance measures, including model-specific metrics. The primary evaluation metric is the Spearman correlation between the predicted similarities and the ground-truth scores.

\paragraph{SICK-NL-STS} In addition to SICKNLPairClassification, we use the SICK-NL annotations for semantic relatedness between sentences. Each sentence pair is assigned a continuous relatedness score ranging from 1 to 5, computed as the average of ten human ratings, which reflects the degree of semantic relatedness between the sentences.

\paragraph{STSBenchmarkMultilingualSTS} This dataset is the machine-translated multilingual extension of the Semantic Textual Similarity benchmark \citep{cer-etal-2017-semeval}. We use the Dutch portion. The dataset was used as provided, without additional preprocessing, as it is already integrated into MTEB.

\section{Appendix B. Models}
Table \ref{tab:models} presents the models used in our experiments, reporting their characteristics along with corresponding citations and Hugging Face URLs.

\begin{table*}[hbp]
\centering
\small
\resizebox{1\textwidth}{!}{ 
\begin{tabular}{@{}p{.41\textwidth} p{.27\textwidth} 
>{\centering\arraybackslash}p{.05\textwidth} 
>{\centering\arraybackslash}p{.05\textwidth} 
>{\centering\arraybackslash}p{.05\textwidth} p{.34\textwidth}@{}}
\toprule
\textbf{Model} & \textbf{Source} & \textbf{Prm} & \textbf{Dim} & \textbf{MaxIn} & \textbf{References} \\ \midrule
% BERT-base & \citet{DBLP:journals/corr/abs-1810-04805} & 109M & 768 & 512 & No & \href{https://huggingface.co/google-bert/bert-base-cased}{\nolinkurl{google-bert/bert-base-cased}} \\
BERTje-base & \citet{devries2019bertje} & 109M & 768 & 512 & \href{https://huggingface.co/GroNLP/bert-base-dutch-cased}{\nolinkurl{GroNLP/bert-base-dutch...}} \\
Tik-to-Tok-base & \citet{remy2023tik} & 116M & 768 & 512 & \href{https://huggingface.co/FremyCompany/olm-bert-oscar-nl-step4}{\nolinkurl{FremyCompany/olm-bert...}} \\
RobBERT-v1-base & \citet{delobelle2020robbert} & 117M & 768 & 512 & \href{https://huggingface.co/pdelobelle/robBERT-base}{\nolinkurl{pdelobelle/robBERT-base}} \\
RobBERT-v2-base & \citet{delobelle2020robbert} & 117M & 1024 & 512 & \href{https://huggingface.co/pdelobelle/robbert-v2-dutch-base}{\nolinkurl{pdelobelle/robbert-v2...}} \\ 
RobBERT-2022-base & \citet{delobelle2022robbert2022} & 119M & 1024 & 512 & \href{https://huggingface.co/DTAI-KULeuven/robbert-2022-dutch-base}{\nolinkurl{DTAI-KULeuven/robbert...}} \\ 
RobBERT-2023-base & \citet{delobelle2023robbert2023conversion} & 124M & 768 & 512 & \href{https://huggingface.co/DTAI-KULeuven/robbert-2023-dutch-base}{\nolinkurl{DTAI-KULeuven/robbert...}} \\
% RoBERTa-base & \citet{DBLP:journals/corr/abs-1907-11692} & 125M & 768 & 512 & \href{https://huggingface.co/FacebookAI/roberta-base}{\nolinkurl{FacebookAI/roberta-base}} \\ 
mBERT-cased-base & \citet{devlin-etal-2019-bert} & 179M & 768 & 512 & \href{https://huggingface.co/google-bert/bert-base-multilingual-cased}{\nolinkurl{google-bert/bert-base...}} \\ 
mDeBERTa-v3-base & \citet{he2023debertav} & 184M & 768 & 512 & \href{https://huggingface.co/microsoft/mdeberta-v3-base}{\nolinkurl{microsoft/mdeberta-v3-base}} \\ 
% DeBERTa-v3-base & \citet{he2021debertav3, he2021deberta} & 190M & 768 & 256 & \href{https://huggingface.co/microsoft/mdeberta-v3-base}{\nolinkurl{microsoft/mdeberta-v3-base}} \\ 
XLM-R-base & \citet{DBLP:journals/corr/abs-1911-02116} & 279M & 768 & 512 & \href{https://huggingface.co/FacebookAI/xlm-roberta-base}{\nolinkurl{FacebookAI/xlm-roberta-base}} \\ 
% DeBERTa-v3-large & \citet{he2021debertav3, he2021deberta} & 304M & 1024 & 256 & \href{https://huggingface.co/microsoft/deberta-v3-large}{\nolinkurl{microsoft/deberta-v3-large}} \\
Tik-to-Tok-large & \citet{remy2023tik} & 345M & 1024 & 512 & \href{https://huggingface.co/FremyCompany/rl-bert-oscar-nl-step4}{\nolinkurl{FremyCompany/rl-bert...}} \\
% BERT-large & \citet{DBLP:journals/corr/abs-1810-04805} & 355M & 1024 & 512 & \href{https://huggingface.co/google-bert/bert-large-cased}{\nolinkurl{google-bert/bert-large-cased}} \\ 
RobBERT-2023-large & \citet{delobelle2023robbert2023conversion} & 355M & 1024 & 512 & \href{https://huggingface.co/FremyCompany/roberta-large-nl-oscar23}{\nolinkurl{FremyCompany/roberta...}} \\
% RoBERTa-large & \citet{DBLP:journals/corr/abs-1907-11692} & 355M & 1024 & 512 & \href{https://huggingface.co/FacebookAI/roberta-large}{\nolinkurl{FacebookAI/roberta-large}} \\ 
XLM-R-large & \citet{DBLP:journals/corr/abs-1911-02116} & 561M & 1024 & 512 & \href{https://huggingface.co/FacebookAI/xlm-roberta-large}{\nolinkurl{FacebookAI/xlm-roberta-large}} \\ 
\midrule

static-similarity-mrl-multilingual-v1 & \citet{reimers-gurevych-2019-sentence} & - & 1024 & - & \href{https://huggingface.co/sentence-transformers/static-similarity-mrl-multilingual-v1}{\nolinkurl{sentence-transformers/sta...}} \\
Granite-Embedding-107m-multilingual & \citet{awasthy2025graniteembeddingmodels} & 107M & 384 & 512 & \href{https://huggingface.co/ibm-granite/granite-embedding-107m-multilingual}{\nolinkurl{ibm-granite/granite...}} \\
multilingual-e5-small & \citet{wang2024multilingual} & 118M & 384 & 512 & \href{https://huggingface.co/intfloat/multilingual-e5-small}{\nolinkurl{intfloat/multilingual-e5}} \\
paraphrase-multilingual-MiniLM-L12-v2 & \citet{reimers-gurevych-2019-sentence} & 118M & 768 & 512 & \href{https://huggingface.co/sentence-transformers/paraphrase-multilingual-MiniLM-L12-v2}{\nolinkurl{sentence-transformers/par...}} \\
potion-multilingual-128M & \citet{minishlab2024model2vec} & 128M & 256 & - & \href{https://huggingface.co/minishlab/potion-multilingual-128M}{\nolinkurl{minishlab/potion...}} \\
multilingual-e5-base & \citet{wang2024multilingual} & 278M & 768 & 512 & \href{https://huggingface.co/intfloat/multilingual-e5-base}{\nolinkurl{intfloat/multilingual-e5}} \\
Granite-Embedding-278m-multilingual & \citet{awasthy2025graniteembeddingmodels} & 278M & 768 & 512 & \href{https://huggingface.co/ibm-granite/granite-embedding-278m-multilingual}{\nolinkurl{ibm-granite/granite...}} \\
paraphrase-multilingual-mpnet-base-v2 & \citet{reimers-gurevych-2019-sentence} & 278M & 768 & 512 & \href{https://huggingface.co/sentence-transformers/paraphrase-multilingual-mpnet-base-v2}{\nolinkurl{sentence-transformers/par...}} \\
Arctic-embed-m-v2.0 & \citet{yu2024arcticembed20multilingualretrieval} & 305M & 768 & 8K & \href{https://huggingface.co/Snowflake/snowflake-arctic-embed-m-v2.0}{\nolinkurl{Snowflake/snowflake...}} \\
gte-multilingual-base & \citet{zhang2024mgte} & 305M & 768 & 8K & \href{https://huggingface.co/Alibaba-NLP/gte-multilingual-base}{\nolinkurl{Alibaba-NLP/gte-multi...}} \\
LaBSE & \citet{feng-etal-2022-language} & 471M & 768 & 512 & \href{https://huggingface.co/sentence-transformers/LaBSE}{\nolinkurl{sentence-transformers/LaBSE}} \\
multilingual-e5-large & \citet{wang2024multilingual} & 560M & 1024 & 512 & \href{https://huggingface.co/intfloat/multilingual-e5-large}{\nolinkurl{intfloat/multilingual-e5...}} \\
Arctic-embed-l-v2.0 & \citet{yu2024arcticembed20multilingualretrieval} & 568M & 1024 & 8K & \href{https://huggingface.co/Snowflake/snowflake-arctic-embed-l-v2.0}{\nolinkurl{Snowflake/snowflake...}} \\
bge-m3 & \citet{bge-m3} & 568M & 1024 & 8K & \href{https://huggingface.co/BAAI/bge-m3}{\nolinkurl{BAAI/bge-m3}} \\
jina-embeddings-v3 & \citet{sturua2024jinaembeddingsv3multilingualembeddingstask} & 572M & 1024 & 8K & \href{https://huggingface.co/jinaai/jina-embeddings-v3}{\nolinkurl{jinaai/jina-embeddings-v3}} \\
% jina-embeddings-v4 & \citet{günther2025jinaembeddingsv4universalembeddingsmultimodal} & 3.75B & 2048 & 127998 & \href{https://huggingface.co/jinaai/jina-embeddings-v4}{\nolinkurl{jinaai/jina-embeddings-v4}} \\
\midrule
KaLM-multilingual-mini-instruct-v1 & \citet{hu2025kalm} & 494M & 896 & 512 & \href{https://huggingface.co/HIT-TMG/KaLM-embedding-multilingual-mini-instruct-v1}{\nolinkurl{HIT-TMG/KaLM-embedding...}} \\
multilingual-e5-large-instruct & \citet{wang2024multilingual} & 560M & 1024 & 512 & \href{https://huggingface.co/intfloat/multilingual-e5-large-instruct}{\nolinkurl{intfloat/multilingual-e5...}} \\
Qwen3-Embedding-0.6B & \citet{qwen3embedding} & 596M & 1024 & 32K & \href{https://huggingface.co/Qwen/Qwen3-Embedding-0.6B}{\nolinkurl{Qwen/Qwen3-Embedding-0.6B}} \\
Qwen3-Embedding-4B & \citet{qwen3embedding} & 4B & 2560 & 32K & \href{https://huggingface.co/Qwen/Qwen3-Embedding-4B}{\nolinkurl{Qwen/Qwen3-Embedding-4B}} \\
% Qwen3-Embedding-8B & \citet{qwen3embedding} & 7.57B & 4096 & 40958 & \href{https://huggingface.co/Qwen/Qwen3-Embedding-8B}{\nolinkurl{Qwen/Qwen3-Embedding-8B}} \\
% BGE-Multilingual-Gemma2 & \citet{bge-m3, bge_embedding} & 9.24B & 3584 & 8190 & \href{https://huggingface.co/BAAI/bge-multilingual-gemma2}{\nolinkurl{BAAI/bge-multilingual-gemma2}} \\
\bottomrule 
\end{tabular}
}
\caption{Self-supervised (top), supervised (middle), and supervised-instruct (bottom) models used in our experiments. \textit{Dim} denotes the embedding dimension, \textit{Prm} the number of model parameters, and \textit{MaxIn} the maximum input length.}
\label{tab:models}
\end{table*}

\section{Appendix C. Results}
Table \ref{tab:all_datasets_with_average} provides a detailed breakdown of MTEB-NL results across datasets.
% \caption{Results on MTEB-NL by dataset for self-supervised (top), supervised (middle), and supervised-instruct (bottom) models.}
% \label{tab:all_datasets_with_average}

% \begin{table*}[tbp]
\begin{sidewaystable*}[tbp]
\small

%\rotatebox{90}{
%\begin{adjustbox}{angle=90}
%\begin{center}
\resizebox{1\textwidth}{!}{ 
\setlength{\tabcolsep}{2pt} % shrink padding inside columns
\begin{tabular}{@{}l*{41}{c}c@{}}
%\toprule
\textbf{Model} &
\rotatebox{90}{DutchBookReviewSentimentClassification} &
\rotatebox{90}{DutchColaClassification} &
\rotatebox{90}{DutchGovernmentBiasClassification} &
\rotatebox{90}{DutchNewsArticlesClassification} &
\rotatebox{90}{DutchSarcasticHeadlinesClassification} &
\rotatebox{90}{IconclassClassification} &
\rotatebox{90}{MassiveIntentClassification} &
\rotatebox{90}{MassiveScenarioClassification} &
\rotatebox{90}{MultiHateClassification} &
\rotatebox{90}{OpenTenderClassification} &
\rotatebox{90}{SIB200Classification} &
\rotatebox{90}{VaccinChatNLClassification} &
\rotatebox{90}{CovidDisinformationNLMultiLabelClassification} &
\rotatebox{90}{MultiEURLEXMultilabelClassification} &
\rotatebox{90}{VABBMultiLabelClassification} &
\rotatebox{90}{SICKNLPairClassification} &
\rotatebox{90}{XLWICNLPairClassification} &
\rotatebox{90}{ArguAna-NL} &
\rotatebox{90}{BelebeleRetrieval} &
\rotatebox{90}{bBSARDNLRetrieval} &
\rotatebox{90}{DutchNewsArticlesRetrieval} &
\rotatebox{90}{LegalQANLRetrieval} &
\rotatebox{90}{NFCorpus-NL} &
\rotatebox{90}{OpenTenderRetrieval} &
\rotatebox{90}{SCIDOCS-NL} &
\rotatebox{90}{SciFact-NL} &
\rotatebox{90}{VABBRetrieval} &
\rotatebox{90}{WebFAQRetrieval} &
\rotatebox{90}{WikipediaRetrievalMultilingual} &
\rotatebox{90}{WikipediaRerankingMultilingual} &
\rotatebox{90}{DutchNewsArticlesClusteringP2P} &
\rotatebox{90}{DutchNewsArticlesClusteringS2S} &
\rotatebox{90}{IconclassClusteringS2S} &
\rotatebox{90}{OpenTenderClusteringP2P} &
\rotatebox{90}{OpenTenderClusteringS2S} &
\rotatebox{90}{SIB200ClusteringS2S} &
\rotatebox{90}{VABBClusteringP2P} &
\rotatebox{90}{VABBClusteringS2S} &
\rotatebox{90}{SICK-NL-STS} &
\rotatebox{90}{STSBenchmarkMultilingualSTS} &
\rotatebox{90}{\textbf{Average}} \\
\midrule
\texttt{BERTje-base} &
65.9 & 60.6 & 64.7 & 57.9 & 65.2 & 40.8 & 39.8 & 45.5 & 60.5 & 22.7 &
60.5 & 30.2 & 45.5 & 25.7 & 40.8 & 72.7 & 69.0 & 21.5 & 51.0 &  1.3 &
3.0  & 17.7 &  8.8 & 17.7 & 11.3 & 19.1 & 12.9 & 49.6 & 53.3 & 72.2 &
36.5 & 17.4 & 35.1 & 24.6 & 31.3 & 11.3 & 19.1 & 15.7 & 53.3 & 53.1 &
36.0 \\
\texttt{Tik-to-Tok-base} &
63.8 & 57.9 & 59.4 & 51.4 & 63.1 & 37.6 & 37.1 & 43.9 & 51.4 & 20.0 &
61.8 & 28.7 & 44.9 & 25.7 & 41.3 & 74.9 & 66.4 & 22.4 & 53.4 &  1.1 &
2.5  &  5.3 &  2.5 & 17.4 &  0.9 &  5.8 &  9.1 & 50.5 & 46.7 & 69.9 &
28.4 & 13.3 & 24.1 &  9.4 &  9.8 & 18.3 & 27.4 & 20.0 & 46.7 & 46.1 &
33.5 \\
\texttt{RobBERT-v1-base } &
60.4 & 57.7 & 60.2 & 50.2 & 60.8 & 40.4 & 40.5 & 47.7 & 53.9 & 20.3 &
57.7 & 28.6 & 43.8 & 22.5 & 35.5 & 71.3 & 69.4 & 15.2 & 42.0 & 2.3 &
6.4 & 18.2 & 4.0 & 19.4 & 1.4 & 6.5 & 24.0 & 15.1 & 42.7 & 72.8 &
31.2 & 9.5 & 19.9 & 11.7 & 14.5 & 27.2 & 29.9 & 22.0 & 51.8 & 52.1 &
34.0 \\
\texttt{RobBERT-v2-base} &
64.8 & 61.6 & 59.2 & 53.6 & 60.4 & 42.1 & 38.8 & 47.8 & 61.0 & 22.3 &
59.2 & 29.7 & 45.7 & 24.2 & 38.4 & 74.9 & 71.1 & 17.4 & 44.6 &  1.4 &
5.7  & 44.6 & 19.7 & 10.3 & 11.6 & 20.5 & 13.6 & 46.1 & 46.6 & 70.9 &
35.8 & 12.2 & 26.6 & 19.9 & 31.8 & 10.3 & 20.5 & 16.8 & 46.6 & 50.4 &
34.9 \\
\texttt{RobBERT-2022-base} &
64.3 & 60.9 & 61.7 & 54.0 & 62.8 & 41.1 & 36.3 & 44.8 & 52.7 & 23.2 &
64.1 & 28.6 & 46.1 & 24.3 & 39.9 & 76.9 & 69.9 & 19.0 & 38.4 &  0.6 &
1.4  & 14.9 &  3.0 & 16.5 &  0.7 &  5.1 & 12.7 & 39.2 & 51.4 & 67.1 &
36.8 & 16.1 & 22.5 & 12.8 & 10.4 & 22.5 & 29.6 & 12.7 & 51.4 & 51.9 &
34.2 \\
\texttt{RobBERT-2023-base} &
64.6 & 59.7 & 61.0 & 51.8 & 64.2 & 44.1 & 36.9 & 45.1 & 54.3 & 21.7 &
61.0 & 32.9 & 45.7 & 25.4 & 40.9 & 75.0 & 70.9 & 22.2 & 54.8 &  2.5 &
3.0  & 54.8 & 15.8 & 17.8 &  9.0 &  9.0 & 15.8 & 54.8 & 64.2 & 64.2 &
31.7 & 11.6 & 28.6 & 10.7 & 10.7 & 16.9 & 29.3 & 17.5 & 46.1 & 50.9 &
35.2 \\

\texttt{mBERT-cased-base} &
54.6 & 54.6 & 68.5 & 38.3 & 65.0 & 47.7 & 36.6 & 42.5 & 53.5 & 15.7 &
60.1 & 25.8 & 45.9 & 24.4 & 35.1 & 72.7 & 64.1 & 11.6 & 33.3 &  1.0 &
2.6  & 16.9 &  2.6 & 19.7 & 11.8 & 25.0 &  9.9 & 32.2 & 50.3 & 65.2 &
19.7 & 14.8 & 32.1 & 23.8 & 23.8 & 11.8 & 24.0 &  6.4 & 50.3 & 48.5 &
31.8 \\
\texttt{mDeBERTa-v3-base} &
51.3 & 68.3 & 18.6 & 23.4 & 62.5 & 25.1 &  8.1 & 12.3 & 50.0 &  2.7 &
23.4 &  4.4 & 45.3 & 13.8 & 14.0 & 73.1 & 58.6 &  1.1 &  3.4 &  0.1 &
0.5  &  0.0 &  0.0 & 12.5 &  4.6 & 11.4 &  0.5 &  2.2 & 40.2 & 40.3 &
 6.6 &  3.3 &  2.3 & 17.5 & 19.6 &  6.1 &  5.2 & 11.4 & 40.2 & 33.8 &
19.9 \\

\texttt{XLM-R-base} &
54.1 & 52.8 & 56.3 & 45.7 & 64.0 & 23.3 & 16.4 & 21.5 & 47.1 &  9.9 &
49.2 &  6.5 & 45.7 & 22.2 & 36.8 & 68.7 & 66.2 & 17.4 &  6.3 &  0.4 &
0.9  &  1.2 &  1.2 & 15.2 &  0.4 &  8.7 &  5.4 &  5.4 &  6.5 & 36.0 &
23.4 &  8.9 & 14.9 &  6.6 &  7.7 & 15.4 & 21.4 & 15.4 & 43.1 & 42.2 &
24.5 \\

\texttt{Tik-to-Tok-large} &
63.8 & 61.3 & 64.4 & 54.8 & 63.5 & 43.5 & 43.2 & 52.6 & 58.8 & 23.1 &
64.4 & 26.7 & 45.3 & 25.3 & 41.6 & 75.7 & 64.0 & 20.1 & 44.3 &  1.4 &
2.5  & 13.3 &  2.5 & 18.2 &  9.7 & 26.2 & 11.8 & 41.6 & 54.6 & 62.4 &
35.3 & 16.7 & 27.5 & 14.4 & 13.2 & 26.2 & 30.6 & 24.4 & 54.6 & 53.7 &
35.0 \\

\texttt{RobBERT-2023-large} &
63.1 & 61.0 & 59.0 & 52.8 & 65.5 & 45.5 & 45.8 & 57.9 & 54.0 & 22.6 &
64.4 & 32.2 & 46.9 & 24.9 & 41.0 & 74.5 & 63.3 & 18.8 & 35.2 &  0.7 &
2.1  & 35.2 & 11.2 & 17.6 &  0.7 &  5.4 & 11.5 & 35.2 & 57.1 & 57.1 &
31.3 & 18.3 & 27.0 & 16.0 & 13.0 & 26.2 & 27.9 & 26.2 & 55.5 & 52.9 &
34.2 \\

\texttt{XLM-R-large} &
53.4 & 56.4 & 59.8 & 49.0 & 59.8 & 31.9 & 18.3 & 22.5 & 45.8 & 13.3 &
55.7 & 15.3 & 43.2 & 24.3 & 38.8 & 69.0 & 59.7 & 20.0 & 25.2 &  1.0 &
1.7  & 25.2 &  9.5 & 16.1 &  1.0 &  5.0 & 18.3 & 22.1 & 39.4 & 22.1 &
25.7 &  3.9 &  9.2 & 10.0 &  7.4 & 13.2 & 23.5 & 22.1 & 43.4 & 36.3 &
25.9 \\

\midrule
\texttt{static-similarity-mrl-multilingual-v1} &
57.4 & 52.8 & 57.6 & 47.0 & 54.2 & 45.6 & 50.8 & 55.6 & 52.0 & 28.0 &
58.8 & 47.8 & 44.0 & 24.3 & 31.1 & 66.7 & 60.6 & 35.7 & 67.7 &  9.8 &
34.5 & 44.8 & 19.3 & 44.8 &  7.7 & 42.3 & 58.2 & 69.0 & 65.9 & 80.5 &
25.9 &  3.7 & 11.3 & 15.1 & 13.2 & 16.4 & 20.9 & 16.2 & 60.6 & 68.4 &
40.1 \\

\texttt{e5-small-v2-t2t} & 60.6 & 54.4 & 61.6 & 50.9 & 59.5 & 49.1 & 53.1 & 61.3 & 54.1 & 32.9 & 69.4 & 37.4 & 42.6 & 29.0 & 43.8 & 89.0 & 60.0 & 33.9 & 85.6 & 11.3 & 50.9 & 63.9 & 18.1 & 31.2 & 8.3 & 37.4 & 58.0 & 57.8 & 83.9 & 85.9 & 28.9 & 15.2 & 21.9 & 19.7 & 17.9 & 33.6 & 29.0 & 26.8 & 71.1 & 77.5 & 46.9 \\

\texttt{e5-small-v2-t2t-nl} & 62.9 & 55.0 & 63.4 & 72.5 & 53.4 & 40.4 & 54.3 & 61.6 & 60.1 & 52.9 & 35.9 & 51.2 & 88.4 & 61.4 & 42.8 & 31.1 & 48.7 & 17.6 & 34.0 & 38.5 & 29.0 & 37.1 & 21.0 & 26.0 & 21.1 & 85.9 & 44.9 & 11.0 & 46.7 & 22.2 & 89.9 & 63.2 & 58.0 & 13.3 & 65.6 & 33.9 & 64.3 & 85.7 & 71.4 & 76.9 & 49.8 \\

\texttt{e5-small-trm} & 62.2 & 55.4 & 62.3 & 55.8 & 68.5 & 47.1 & 54.5 & 63.6 & 54.2 & 38.4 & 71.5 & 42.9 & 49.6 & 31.1 & 49.9 & 88.5 & 64.4 & 39.9 & 92.9 & 12.9 & 60.9 & 72.6 & 28.0 & 34.2 & 8.9 & 60.9 & 65.7 & 71.3 & 88.7 & 87.3 & 43.3 & 24.4 & 20.8 & 22.9 & 15.0 & 35.3 & 34.6 & 29.3 & 71.9 & 76.6 & 51.4 \\

\texttt{e5-small-trm-nl} & 64.5 & 56.3 & 66.4 & 75.5 & 54.4 & 45.2 & 54.9 & 62.8 & 66.5 & 57.5 & 42.1 & 51.8 & 88.1 & 64.0 & 49.0 & 32.6 & 52.4 & 25.8 & 40.3 & 37.1 & 34.6 & 41.6 & 24.0 & 32.0 & 22.6 & 87.1 & 46.2 & 13.4 & 64.6 & 29.2 & 92.4 & 72.5 & 66.6 & 12.7 & 71.4 & 41.6 & 73.2 & 88.7 & 72.2 & 77.0 & 53.8 \\

\texttt{granite-embedding-107m-multilingual} &
54.4 & 54.2 & 59.5 & 52.2 & 61.1 & 49.9 & 50.2 & 60.5 & 55.7 & 37.3 &
71.8 & 39.8 & 47.5 & 28.4 & 49.4 & 77.2 & 62.9 & 45.5 & 84.7 & 9.5 &
61.7 & 52.6 & 23.5 & 37.9 & 13.9 & 58.9 & 23.5 & 85.1 & 85.1 & 84.7 &
36.9 & 19.7 & 23.0 & 30.7 & 17.1 & 37.6 & 41.9 & 31.2 & 64.9 & 71.9 &
49.4 \\
\texttt{e5-base-v2-t2t}  & 61.6 & 54.8 & 62.2 & 51.4 & 59.5 & 48.0 & 54.9 & 62.9 & 55.6 & 33.5 & 70.5 & 38.5 & 44.7 & 28.8 & 47.4 & 84.9 & 61.7 & 39.8 & 88.5 & 11.9 & 53.2 & 63.4 & 18.6 & 28.7 & 7.6 & 43.3 & 58.2 & 57.7 & 84.0 & 85.6 & 35.3 & 16.5 & 21.3 & 20.6 & 17.6 & 35.4 & 31.8 & 25.1 & 69.2 & 77.3 & 47.8 \\

\texttt{e5-base-v2-t2t-nl} & 60.9 & 53.4 & 62.4 & 67.2 & 54.5 & 37.1 & 54.8 & 61.0 & 58.9 & 52.4 & 36.3 & 47.9 & 84.8 & 60.1 & 46.1 & 29.4 & 49.0 & 16.9 & 36.4 & 33.8 & 26.0 & 37.9 & 17.8 & 25.7 & 20.8 & 83.9 & 43.3 & 10.2 & 43.4 & 20.0 & 87.7 & 57.5 & 50.9 & 12.6 & 56.2 & 31.8 & 60.7 & 82.7 & 67.9 & 70.6 & 47.8 \\

\texttt{multilingual-e5-small} &
62.2 & 55.4 & 62.3 & 55.8 & 68.5 & 46.8 & 54.5 & 63.6 & 54.2 & 38.4 &
71.3 & 42.9 & 49.5 & 31.1 & 49.9 & 88.5 & 64.4 & 39.9 & 92.8 & 12.9 &
60.9 & 72.6 & 27.9 & 34.4 &  8.9 & 60.9 & 65.8 & 71.3 & 88.7 & 87.1 &
43.5 & 24.2 & 20.7 & 22.5 & 15.1 & 35.3 & 36.7 & 27.9 & 71.9 & 76.5 &
51.4 \\
\texttt{paraphrase-multilingual-MiniLM-L12-v2} &
57.7 & 52.9 & 58.3 & 47.7 & 58.0 & 47.7 & 57.0 & 64.8 & 54.7 & 37.7 &
70.7 & 52.9 & 46.6 & 25.1 & 42.5 & 92.1 & 64.4 & 32.9 & 81.0 &  9.3 &
43.2 & 37.1 & 16.3 & 37.1 &  9.6 & 33.8 & 47.2 & 43.2 & 71.1 & 80.6 &
30.2 & 21.5 & 23.9 & 26.8 & 26.8 & 23.9 & 34.9 & 33.5 & 73.0 & 79.5 &
46.3 \\

\texttt{RobBERT-2023-base-ft} & 73.5 & 53.9 & 65.3 & 72.1 & 63.4 & 54.2 & 53.9 & 65.3 & 53.9 & 43.8 & 72.1 & 44.2 & 49.0 & 34.1 & 50.6 & 77.0 & 68.5 & 44.4 & 88.8 & 16.4 & 60.4 & 70.8 & 22.0 & 37.0 & 9.9 & 47.9 & 69.0 & 69.0 & 85.8 & 85.7 & 38.1 & 20.3 & 27.5 & 29.3 & 34.8 & 41.0 & 38.8 & 34.9 & 65.9 & 71.1 & 52.0 \\

\texttt{e5-base-trm} & 65.8 & 56.8 & 61.7 & 56.4 & 68.7 & 51.3 & 58.3 & 66.4 & 55.1 & 38.0 & 71.1 & 47.9 & 50.2 & 32.0 & 50.9 & 88.3 & 65.2 & 45.1 & 93.7 & 17.7 & 67.1 & 73.9 & 27.1 & 33.0 & 12.4 & 66.9 & 67.8 & 74.6 & 89.9 & 88.3 & 41.6 & 24.4 & 22.5 & 20.8 & 15.1 & 34.3 & 35.1 & 30.7 & 72.6 & 77.2 & 52.9 \\

\texttt{e5-base-trm-nl} & 69.1 & 60.0 & 68.5 & 74.2 & 55.2 & 49.6 & 56.1 & 61.8 & 66.3 & 56.8 & 43.3 & 54.0 & 89.9 & 66.8 & 49.3 & 34.5 & 53.9 & 30.5 & 38.8 & 40.2 & 36.0 & 42.1 & 27.3 & 35.0 & 24.6 & 87.5 & 46.4 & 14.3 & 67.0 & 28.0 & 93.8 & 74.3 & 66.3 & 19.7 & 67.4 & 39.9 & 72.9 & 89.1 & 73.7 & 77.9 & 55.0 \\

\texttt{potion-multilingual-128M} &
55.9 & 54.0 & 61.6 & 56.0 & 57.1 & 45.5 & 50.1 & 62.1 & 51.1 & 33.0 &
61.5 & 34.2 & 47.9 & 26.5 & 45.6 & 59.2 & 61.6 & 36.7 & 72.5 & 12.7 &
25.4 & 36.5 & 16.4 & 28.6 &  7.0 & 41.4 & 51.2 & 34.1 & 65.6 & 80.3 &
39.4 & 14.1 & 20.1 & 22.2 & 17.7 & 29.7 & 40.2 & 25.5 & 57.8 & 66.2 &
42.6 \\
\texttt{multilingual-e5-base} &
65.9 & 56.8 & 61.6 & 56.4 & 68.8 & 51.4 & 58.3 & 66.4 & 55.2 & 38.0 &
71.1 & 47.9 & 50.1 & 32.0 & 51.0 & 88.3 & 65.2 & 45.1 & 93.7 & 17.9 &
67.1 & 74.4 & 27.1 & 33.1 & 12.4 & 66.8 & 67.8 & 74.6 & 89.9 & 88.4 &
41.2 & 24.2 & 21.9 & 20.7 & 14.8 & 36.2 & 34.3 & 28.3 & 72.6 & 77.2 &
52.8 \\
\texttt{granite-embedding-278m-multilingual} &
54.5 & 54.5 & 61.9 & 52.0 & 62.3 & 48.2 & 51.3 & 61.9 & 56.2 & 40.2 &
72.3 & 41.1 & 48.6 & 29.8 & 50.1 & 78.0 & 64.0 & 48.2 & 85.6 & 10.0 &
66.9 & 53.8 & 24.3 & 40.7 & 14.3 & 60.2 & 24.3 & 87.0 & 86.3 & 85.6 &
38.3 & 20.4 & 22.7 & 33.1 & 18.2 & 38.4 & 39.5 & 31.9 & 65.1 & 72.7 &
50.5 \\
\texttt{paraphrase-multilingual-mpnet-base-v2} &
59.6 & 53.6 & 60.0 & 50.2 & 61.1 & 52.6 & 60.6 & 70.0 & 58.7 & 42.0 &
74.2 & 54.9 & 48.7 & 28.2 & 44.5 & 95.5 & 68.5 & 34.3 & 83.9 &  9.6 &
53.9 & 41.4 & 18.5 & 41.4 & 11.2 & 42.2 & 42.3 & 50.7 & 76.7 & 82.3 &
31.2 & 21.4 & 26.9 & 30.7 & 27.6 & 26.9 & 35.2 & 33.1 & 75.3 & 83.4 &
49.2 \\

\texttt{Arctic-embed-m-v2.0} &
56.2 & 53.6 & 62.0 & 52.5 & 61.4 & 51.6 & 54.1 & 62.0 & 53.6 & 36.3 &
67.1 & 42.5 & 49.8 & 27.8 & 50.3 & 73.1 & 60.2 & 46.4 & 84.2 & 13.5 &
61.5 & 70.6 & 26.2 & 31.9 & 11.9 & 66.8 & 26.2 & 84.2 & 86.8 & 86.2 &
33.5 & 18.7 & 24.9 & 19.3 & 17.6 & 32.7 & 37.0 & 28.1 & 63.7 & 66.0 &
49.1 \\

\texttt{gte-multilingual-base} 
& 76.7 & 55.5 & 61.5 & 53.0 & 65.8 & 53.9 
& 59.4 & 68.9 & 55.7 & 42.6 & 67.7 & 48.8 
& 49.1 & 11.6 & 52.3 & 92.9 & 62.7 
& 52.8 & 89.2 & 20.1 & 80.9 & 64.9 & 29.5 
& 42.3 & 15.8 & 64.3 & 73.7 & 64.3 & 84.0 & 82.3 
& 35.3 & 29.8 & 26.4 & 33.6 & 28.2 & 25.8 
& 38.0 & 33.3 & 75.8 & 81.3 & 53.8 \\
\texttt{e5-large-v2-t2t} & 70.0 & 54.6 & 60.5 & 53.2 & 61.4 & 47.6 & 55.6 & 62.1 & 55.6 & 35.3 & 72.0 & 40.2 & 45.0 & 31.0 & 48.3 & 88.9 & 62.5 & 34.1 & 89.7 & 16.9 & 58.0 & 66.7 & 22.7 & 32.6 & 10.8 & 52.8 & 64.7 & 62.7 & 86.9 & 86.6 & 36.6 & 15.9 & 20.7 & 21.0 & 18.8 & 35.5 & 30.0 & 25.8 & 70.9 & 77.0 & 49.5 \\
\texttt{e5-large-v2-t2t-nl} & 79.0 & 57.0 & 64.4 & 72.5 & 54.9 & 44.1 & 55.1 & 61.6 & 63.9 & 56.5 & 38.6 & 50.9 & 84.3 & 65.0 & 48.1 & 34.6 & 51.1 & 20.9 & 37.5 & 37.8 & 29.5 & 44.9 & 15.9 & 26.2 & 22.3 & 87.1 & 43.6 & 12.1 & 55.7 & 23.2 & 91.0 & 67.2 & 57.2 & 18.5 & 60.2 & 38.1 & 66.5 & 87.7 & 69.4 & 75.1 & 51.7 \\
\texttt{RobBERT-2023-large-ft} & 81.8 & 58.0 & 66.9 & 68.8 & 62.8 & 53.4 & 55.2 & 66.9 & 56.5 & 39.4 & 68.8 & 48.6 & 52.3 & 33.5 & 49.7 & 67.6 & 69.7 & 44.3 & 87.9 & 16.1 & 50.3 & 58.6 & 21.6 & 36.0 & 9.2 & 47.8 & 64.6 & 64.6 & 82.9 & 82.9 & 38.3 & 27.1 & 28.0 & 27.1 & 28.7 & 35.6 & 38.6 & 29.4 & 65.2 & 76.1 & 51.0 \\

\texttt{e5-large-trm} & 65.6 & 56.8 & 63.1 & 57.9 & 72.8 & 51.1 & 62.0 & 69.4 & 59.0 & 40.9 & 73.4 & 50.6 & 49.0 & 34.1 & 52.9 & 93.3 & 67.3 & 48.9 & 95.3 & 23.8 & 74.6 & 77.5 & 29.8 & 37.8 & 13.1 & 68.4 & 70.3 & 77.2 & 91.9 & 90.3 & 42.4 & 24.9 & 22.9 & 23.1 & 17.2 & 36.5 & 33.6 & 29.2 & 76.9 & 80.8 & 55.1 \\

\texttt{e5-large-trm-nl} & 76.6 & 62.5 & 70.6 & 76.2 & 60.7 & 51.5 & 55.9 & 61.3 & 74.0 & 58.2 & 45.9 & 53.2 & 95.3 & 67.5 & 51.7 & 36.3 & 56.1 & 30.7 & 40.0 & 42.3 & 35.5 & 45.6 & 29.5 & 36.1 & 25.6 & 87.2 & 47.1 & 16.0 & 63.9 & 30.9 & 93.1 & 74.3 & 71.1 & 22.7 & 71.5 & 42.4 & 76.2 & 88.8 & 75.8 & 80.6 & 57.0 \\

\texttt{LaBSE} &
60.9 & 55.7 & 60.4 & 53.3 & 59.4 & 52.9 & 56.6 & 63.2 & 56.2 & 35.7 &
58.0 & 43.2 & 49.6 & 26.4 & 48.3 & 82.5 & 69.6 & 39.2 & 74.9 &  3.5 &
45.9 & 30.7 & 15.5 & 25.0 &  6.3 & 39.0 & 38.9 & 39.0 & 65.3 & 80.5 &
37.5 & 18.2 & 25.9 & 20.6 & 16.9 & 23.2 & 34.6 & 23.2 & 67.8 & 70.2 &
44.5 \\
\texttt{multilingual-e5-large} &65.6 & 56.8 & 63.1 & 57.8 & 72.8 & 51.3 & 62.0 & 69.4 & 59.1 & 40.8 &
73.4 & 50.7 & 49.2 & 34.1 & 52.8 & 93.3 & 67.3 & 48.9 & 95.2 & 23.8 &
74.6 & 77.5 & 29.8 & 37.9 & 13.1 & 68.4 & 70.3 & 77.2 & 91.8 & 90.3 &
41.9 & 24.7 & 23.2 & 23.0 & 16.9 & 39.6 & 34.8 & 31.7 & 76.9 & 80.8 &
55.3 \\
\texttt{Arctic-embed-l-v2.0} &
64.9 & 55.2 & 62.6 & 52.9 & 65.9 & 56.3 & 61.9 & 71.2 & 54.1 & 43.5 &
71.1 & 51.3 & 50.8 & 31.5 & 53.2 & 81.7 & 66.8 & 53.2 & 90.9 & 24.6 &
72.8 & 80.3 & 29.3 & 37.9 & 15.9 & 68.1 & 73.0 & 72.8 & 89.4 & 88.2 &
33.9 & 26.5 & 24.0 & 24.7 & 23.7 & 35.0 & 38.6 & 32.2 & 69.8 & 73.6 &
54.3 \\

\texttt{bge-m3} &
77.7 & 56.2 & 61.0 & 55.7 & 65.3 & 51.3 & 66.0 & 73.2 & 57.0 & 41.1 &
71.5 & 52.4 & 50.3 & 30.7 & 51.6 & 92.2 & 64.4 & 52.1 & 93.8 & 24.1 &
81.7 & 81.2 & 29.0 & 39.1 & 14.8 & 62.9 & 75.0 & 76.7 & 90.1 & 88.7 &
38.3 & 22.1 & 23.2 & 25.6 & 22.4 & 34.6 & 37.2 & 29.7 & 76.3 & 79.8 &
55.4 \\

\texttt{jina-embeddings-v3} &
90.9 & 54.5 & 61.7 & 52.2 & 62.5 & 53.1 & 67.4 & 80.8 & 56.6 & 40.0 &
74.9 & 45.9 & 49.0 & 15.7 & 52.1 & 87.4 & 66.1 & 36.1 & 93.1 & 24.9 &
76.5 & 73.1 & 32.6 & 41.7 & 17.1 & 69.6 & 74.2 & 79.8 & 90.3 & 78.5 &
45.3 & 31.7 & 30.3 & 36.0 & 34.3 & 43.8 & 51.5 & 38.2 & 83.1 & 86.5 &
57.0 \\

\midrule
\texttt{KaLM-multilingual-mini-instruct-v1} &
85.1 & 54.2 & 77.5 & 57.5 & 66.0 & 48.6 & 58.6 & 73.7 & 57.5 & 39.4 &
77.5 & 38.4 & 48.9 & 34.2 & 54.5 & 86.6 & 66.1 & 53.0 & 91.1 & 21.9 &
67.9 & 60.9 & 28.4 & 34.1 & 14.9 & 61.8 & 68.1 & 91.0 & 87.0 & 85.5 &
41.8 & 30.9 & 50.1 & 37.3 & 52.4 & 31.1 & 39.6 & 28.8 & 70.8 & 75.9 &
54.9 \\
\texttt{multilingual-e5-large-instruct} &
91.3 & 53.8 & 82.7 & 57.7 & 67.0 & 50.5 & 65.6 & 58.7 & 67.0 & 49.0 &
50.5 & 51.0 & 50.9 & 35.1 & 56.1 & 94.3 & 66.9 & 49.0 & 16.9 & 69.5 &
32.1 & 93.9 & 75.7 & 82.3 & 33.4 & 79.9 & 39.4 & 75.2 & 89.9 & 88.2 &
50.9 & 30.8 & 16.9 & 69.5 & 75.7 & 39.4 & 75.2 & 83.7 & 78.9 & 83.7 &
60.6 \\
\texttt{Qwen3-Embedding-0.6B} &
92.5 & 56.1 & 80.8 & 55.0 & 64.9 & 47.6 & 64.1 & 73.8 & 55.0 & 45.4 &
80.8 & 32.2 & 50.4 & 34.4 & 54.4 & 85.9 & 61.3 & 56.5 & 91.7 & 19.9 &
60.7 & 69.5 & 27.2 & 46.8 & 17.7 & 67.4 & 71.6 & 80.5 & 86.7 & 85.1 &
39.3 & 26.3 & 51.7 & 37.3 & 53.7 & 32.8 & 39.4 & 29.0 & 73.1 & 80.5 &
56.1 \\
\texttt{Qwen3-Embedding-4B} &
96.4 & 60.5 & 82.5 & 62.0 & 74.2 & 58.3 & 78.1 & 84.9 & 61.8 & 54.3 &
82.5 & 39.0 & 55.0 & 38.1 & 58.4 & 94.9 & 66.3 & 59.3 & 95.5 & 38.5 &
75.8 & 79.2 & 35.1 & 50.0 & 23.4 & 75.9 & 77.6 & 88.9 & 89.2 & 87.1 &
47.1 & 35.3 & 53.2 & 45.6 & 59.5 & 45.7 & 52.9 & 37.6 & 82.8 & 88.9 &
63.6 \\

\bottomrule
\end{tabular}

}
%}
%\end{center}

%\end{adjustbox}
\caption{Results on MTEB-NL by dataset for self-supervised (top), supervised (middle), and supervised-instruct (bottom) models.}
\label{tab:all_datasets_with_average}
% \end{table*}
\end{sidewaystable*}

\section{Appendix D. Prompts for Synthetic Data}
Tables \ref{tab:prompt-1}, \ref{tab:prompt-2}, \ref{tab:prompt-3}, \ref{tab:prompt-4} and \ref{tab:prompt-5} show the prompt templates and parameters used for the 5 different categories of synthetic data. Templates are based on \citet{wang-etal-2024-improving-text}.

\begin{table*}[ht]
%\centering
\small
\begin{tabular}{p{16cm}}
\toprule
\textbf{Prompt}:\\  
"To train a SOTA Dutch retrieval model we want to generate high quality synthetic data in Dutch, for the following task: \{task\} \\
Your mission is to write one text retrieval example for this task in JSON format. The JSON object must contain the following keys: \\
- "user-query": a string, a random user search query. \\
- "positive-document": a string, a relevant document for the user query.\\
- "hard-negative-document": a string, a hard negative document that only appears relevant to the query.\\ 
Please adhere to the following guidelines:\\
- All generated texts should be in Dutch. \\
- The queries and documents should be about \{topics\}. \\
- The "user-query" should be \{query-type\}, \{query-length\}, \{clarity\} \{lexical-overlap\}. \\
- All documents must be created independent of the query. Avoid copying the query verbatim. It’s acceptable if some parts of the "positive-document" are not topically related to the query. \\
- All documents should be at least \{num-words\} words long. \\
- The "hard-negative-document" contains some useful information, but it should be less useful or comprehensive compared to the "positive-document". \\
- Do not provide any explanation in any document on why it is relevant or not relevant to the query. \\
- Both the query and documents require a \{difficulty\} level education to understand. \\
- \{local-flag\} \\
Your output must always be a JSON object only, do not explain yourself or output anything else. Be creative!" \\

\midrule
\textbf{Parameters:}\\
-\texttt{task}: \\
 \hspace{.3cm}   - Given a question, retrieve documents that can help to answer the question.\\
 \hspace{.3cm}   - Given a query, retrieve documents that fulfill the informational needs of the query; e.g. explain, expand, analyze, etc.\\
  \hspace{.3cm}  - Given a claim, retrieve documents that support or refute it.\\

- \texttt{topics}: Two topics ($T_{1}$ and $T_{2}|T_{1}$) sampled from our distribution (c.f. \ref{synth-data})\\
- \texttt{query-type} $\in$ ["Extremely long-tail", "Long-tail", "Common"]\\
- \texttt{query-length} $\in$ ["Less than 7 words", "7 to 17 words", "At least 12 words"]\\
- \texttt{clarity} $\in$ ["Clear", "Understandable with some effort", "Ambiguous"]\\
- \texttt{lexical-overlap} $\in$ ["", "and have Minimum lexical overlap with the "positive document"]\\
- \texttt{num-words} $\in$ [50, 100, 200, 300, 400, 500]\\
- \texttt{difficulty} $\in$ ["Layman", "High school", "Bachelor's degree", "Master's degree or higher"]\\
- \texttt{local-flag} : "If possible, try to generate the example in the Flemish or Dutch context (e.g. including Flemish/Dutch entities, events, etc.)."\\
\bottomrule
\end{tabular}
\caption{Prompt template and parameters for the short-long data category; i.e. retrieval. The local flag is added randomly in 1/3 of samples.}
\label{tab:prompt-1}
\end{table*}

\begin{table*}[ht]
%\centering
\small
\begin{tabular}{p{16cm}}
\toprule
\textbf{Prompt}:\\  
You have been assigned a text matching task: \{task\} \\
Your mission is to write one example for this task in JSON format. The JSON object must contain the following keys: \\
- "input": a string, a random input specified by the task. \\
- "positive-document": a string, a relevant document for the "input" according to the task. \\
- "hard-negative-document": a hard negative document that ONLY APPEARS relevant to the "input" (according to the task). \\

Please adhere to the following guidelines: \\
- The values of all fields should be in Dutch. \\
- "input", "positive-document" and "hard-negative-document" should be  very short (a sentence or a phrase). If compatible with the task, they should be about \{topics\}. \\
- Avoid substantial word overlaps between "input" and "positive-document". Otherwise the task would be too easy. \\
- The "input" and "positive-document" should be generated independent of each other. \\
- \{local-flag\} \\
Your output must always be a JSON object only, do not explain yourself or output anything else. Be creative!\\

\midrule
\textbf{Parameters:}\\
-\texttt{task}: \\
 \hspace{.3cm} - Given the title of a forum post (e.g. from StackExchange, Reddit, etc.), find post titles that belong to the same forum category/topic.\\
 \hspace{.3cm} - Given a news headline, find others that belong to the same category/topic.\\
 \hspace{.3cm}  - Given a premise, find entailing hypotheses.\\
 \hspace{.3cm}  - Given the title of a scientific paper, find titles that belong to the same scientific disciplines/categories/topics.\\

- \texttt{topics}: Two topics ($T_{1}$ and $T_{2}|T_{1}$) sampled from our distribution (c.f. \ref{synth-data})\\
- \texttt{local-flag} : "If possible, try to generate the example in the Flemish or Dutch context (e.g. including Flemish/Dutch entities, events, etc.)."\\
\bottomrule
\end{tabular}
\caption{Prompt template and parameters for the short-short data category. The local flag is added randomly in 1/3 of samples.}
\label{tab:prompt-2}
\end{table*}

\begin{table*}[ht]
%\centering
\small
\begin{tabular}{p{16cm}}
\toprule
\textbf{Prompt}:\\  
You have been assigned a text matching task: \{task\} \\
Your mission is to write one example for this task in JSON format. The JSON object must contain the following keys: \\
- "input": a string, a random input specified by the task. \\
- "positive-document": a string, a relevant document for the "input" according to the task.\\
- "hard-negative-document": a hard negative document that ONLY APPEARS relevant to the "input" (according to the task). \\

Please adhere to the following guidelines:\\
- The values of all fields should be in Dutch.\\
- "input", "positive-document" and "hard-negative-document" should be long documents (at least 300 words). If compatible with the task, they should be about \{topics\}. \\
- Avoid substantial word overlaps between "input" and "positive-document". Otherwise the task would be too easy.\\
- The "input" and "positive-document" should be generated independent of each other. \\
- \{local-flag\}\\
Your output must always be a JSON object only, do not explain yourself or output anything else. Be creative!\\

\midrule
\textbf{Parameters:}\\
-\texttt{task}: \\
 \hspace{.3cm} - Given a forum post (e.g. from StackExchange, Reddit, etc.), find posts that belong to the same forum category/topic.\\
 \hspace{.3cm} - Given a news article, find others that belong to the same category/topic.\\
 \hspace{.3cm}  - Given a document that supports a debatable argument, find documents that contain opposite arguments.\\
 \hspace{.3cm}  - Given a scientific abstract, find abstracts that belong to the same scientific disciplines/categories/topics.\\

- \texttt{topics}: Two topics ($T_{1}$ and $T_{2}|T_{1}$) sampled from our distribution (c.f. \ref{synth-data})\\
- \texttt{local-flag} : "If possible, try to generate the example in the Flemish or Dutch context (e.g. including Flemish/Dutch entities, events, etc.)."\\
\bottomrule
\end{tabular}
\caption{Prompt template and parameters for the long-long data category. The local flag is added randomly in 1/3 of samples.}
\label{tab:prompt-3}
\end{table*}

\begin{table*}[ht]
%\centering
\small
\begin{tabular}{p{16cm}}
\toprule
\textbf{Prompt}:\\  
You have been assigned a text classification task: \{task\} \\
Your mission is to write one text classification example for this task in JSON format. The JSON object must contain the following keys:\\
- "input-text": a string, the input text specified by the classification task.\\
- "label": a string, the correct label of the input text.\\
- "misleading-label": a string, an incorrect label that is related to the task.\\

Please adhere to the following guidelines:\\
- The "input-text" should be \{num-words\} words long, and diverse in expression. If compatible with the task, it should be about \{topics\}.\\
- The "misleading-label" must be a valid label for the given task, but not as appropriate as the "label" for the "input-text".\\
- The values for all fields should be in Dutch.\\
- Avoid including the values of the "label" and "misleading-label" fields in the "input-text", that would make the task too easy.\\
- The "input-text" is \{clarity\} and requires \{difficulty\} level education to comprehend.\\
- \{local-flag\}\\
Your output must always be a JSON object only, do not explain yourself or output anything else. Be creative!\\

\midrule
\textbf{Parameters:}\\
-\texttt{task}: \\
 \hspace{.3cm} - Identifying the polarity of a user opinion, review or post\\
 \hspace{.3cm} - Identifying the positivity level of a user opinion, review\\
 \hspace{.3cm} - Identifying the intent or scenario of a user utterance, input, query or command\\
 \hspace{.3cm} - Identifying the emotion of a user opinion, review or post\\
 \hspace{.3cm} - Identifying the toxicity of a user opinion, review or post\\
 \hspace{.3cm} - Identifying the topic of a text like question, query, news, forum post, etc.\\ 
 \hspace{.3cm} - Identifying the category of a text like news headline or summary, article title or abstract, forum post, etc.\\

- \texttt{topics}: Two topics ($T_{1}$ and $T_{2}|T_{1}$) sampled from our distribution (c.f. \ref{synth-data})\\
- \texttt{query-length} $\in$ ["less than 10", "at least 10", "at least 50", "at least 100", "at least 200"]\\
- \texttt{difficulty}  $\in$ ["Layman", "High school", "Bachelor's degree", "Master's degree or higher"]\\
- \texttt{clarity}  $\in$ ["Clear", "Understandable with some effort", "Ambiguous"]\\
- \texttt{local-flag} : "If possible, try to generate the example in the Flemish or Dutch context (e.g. including Flemish/Dutch entities, events, etc.)."\\
\bottomrule
\end{tabular}
\caption{Prompt template and parameters for the long-short data category; i.e. classification. The local flag is added randomly in 1/3 of samples.}
\label{tab:prompt-4}
\end{table*}

\begin{table*}[ht]
%\centering
\small
\begin{tabular}{p{16cm}}
\toprule
\textbf{Prompt}:\\  
We want to generate high-quality synthetic data for semantic textual similarity (STS) in Dutch.\\
Your mission is to write a \{unit\} triple with varying semantic similarity scores in JSON format. The semantic similarity score ranges from 1 to 5, with 1 denoting least similar and 5 denoting most similar.\\

Please adhere to the following guidelines:\\
- The keys in JSON are "S1", "S2", and "S3", the values are all strings in Dutch. Do not add any other keys.\\
- The \{unit\}s should be about \{topics\}. \{local-flag\}\\
- There should be some word overlaps between all three \{unit\}s.\\
- The similarity score between S1 and S2 should be \{high-score\}.\\
- The similarity score between S1 and S3 should be \{low-score\}.\\
- The \{unit\}s require \{difficulty\} level education to understand.\\

Your output must always be a JSON object only with three keys "S1", "S2" and "S3", do not explain yourself or output anything else. Be creative!\\

\midrule
\textbf{Parameters:}\\
- \texttt{unit} $\in$ ["sentence", "phrase", "passage"]\\
- \texttt{topics}: Two topics ($T_{1}$ and $T_{2}|T_{1}$) sampled from our distribution (c.f. \ref{synth-data})\\
- \texttt{local-flag} : "If possible, try to generate them in the Flemish or Dutch context (e.g. including Flemish/Dutch entities, events, etc.)."\\
- \texttt{high-score}  $\in$ [4, 4.5, 5] \\
- \texttt{hlow-score}  $\in$ [2.5, 3, 3.5]\\
- \texttt{difficulty}  $\in$ ["Layman", "High school", "Bachelor's degree", "Master's degree or higher"]\\

\bottomrule
\end{tabular}
\caption{Prompt template and parameters for the STS data category; i.e. semantic text similarity. The local flag is added randomly in 1/3 of samples.}
\label{tab:prompt-5}
\end{table*}

\end{document}